\documentclass[10pt,twocolumn,letterpaper]{article}

\usepackage{cvpr}
\usepackage{times}
\usepackage{epsfig}
\usepackage{epstopdf}
\usepackage{float}
\usepackage{graphicx}
\usepackage{amsmath}
\usepackage{amsthm}
\usepackage{amssymb}
\usepackage{booktabs}
\usepackage[lined,boxed,ruled]{algorithm2e}

\newcommand{\R}[1]{\mathbb{R}^{#1}}
\newcommand{\RR}[2]{\mathbb{R}^{#1 \times #2}}

\newcommand{\pr}{\mbox{Pr}}

\DeclareMathOperator*{\argmax}{arg\!\max}
\DeclareMathOperator*{\argmin}{arg\!\min}

\newcommand{\refEq}[1]{(\ref{#1})}
\newcommand{\refFig}[1]{Figure~\ref{#1}}

\newcommand{\refSec}[1]{Section~\ref{#1}}

\newcommand{\refAlg}[1]{Algorithm~\ref{#1}}
\newcommand{\refTab}[1]{Table~\ref{#1}}

\def\bfc{{\boldsymbol{c}}}

\def\bfw{{\boldsymbol{w}}}

\def\bfB{{\boldsymbol{B}}}
\def\bfC{{\boldsymbol{C}}}

\def\bfI{{\boldsymbol{I}}}

\def\bfR{{\boldsymbol{R}}}
\def\bfS{{\boldsymbol{S}}}
\def\bfT{{\boldsymbol{T}}}

\def\bfW{{\boldsymbol{W}}}

\def\bfY{{\boldsymbol{Y}}}

\def\bfone{{\boldsymbol{1}}}

\usepackage[pagebackref=true,breaklinks=true,letterpaper=true,colorlinks,bookmarks=false]{hyperref}

\cvprfinalcopy 

\def\cvprPaperID{1648} 

\begin{document}

\title{\vspace{-0.5em} Sparseness Meets Deepness: 3D Human Pose Estimation from Monocular Video}
\author{Xiaowei Zhou$^{\dag}$\thanks{The first two authors contributed equally to this work.}, Menglong Zhu$^{\dag*}$, Spyridon Leonardos$^\dag$, Konstantinos G.\ Derpanis$^{\ddag}$, Kostas Daniilidis$^\dag$ \\[0ex]
$^\dag$ University of Pennsylvania \hspace{2em} $^\ddag$ Ryerson University\\
}

\maketitle

\begin{abstract}

This paper addresses the challenge of 3D full-body human pose estimation from a monocular image sequence. Here, two cases are considered: (i) the image locations of the human joints are provided and (ii) the image locations of joints are unknown. In the former case, a novel approach is introduced that integrates a sparsity-driven 3D geometric prior and temporal smoothness. In the latter case, the former case is extended by treating the image locations of the joints as latent variables to take into account considerable uncertainties in 2D joint locations. A deep fully convolutional network is trained to predict the uncertainty maps of the 2D joint locations. The 3D pose estimates are realized via an Expectation-Maximization algorithm over the entire sequence, where it is shown that the 2D joint location uncertainties can be conveniently marginalized out during inference. Empirical evaluation on the Human3.6M dataset shows that the proposed approaches achieve greater 3D pose estimation accuracy over  state-of-the-art baselines.  Further, the proposed approach outperforms a publicly available 2D pose estimation baseline on the challenging PennAction dataset.

\end{abstract}

\section{Introduction}


\begin{figure}[t]
   \includegraphics[width=\linewidth]{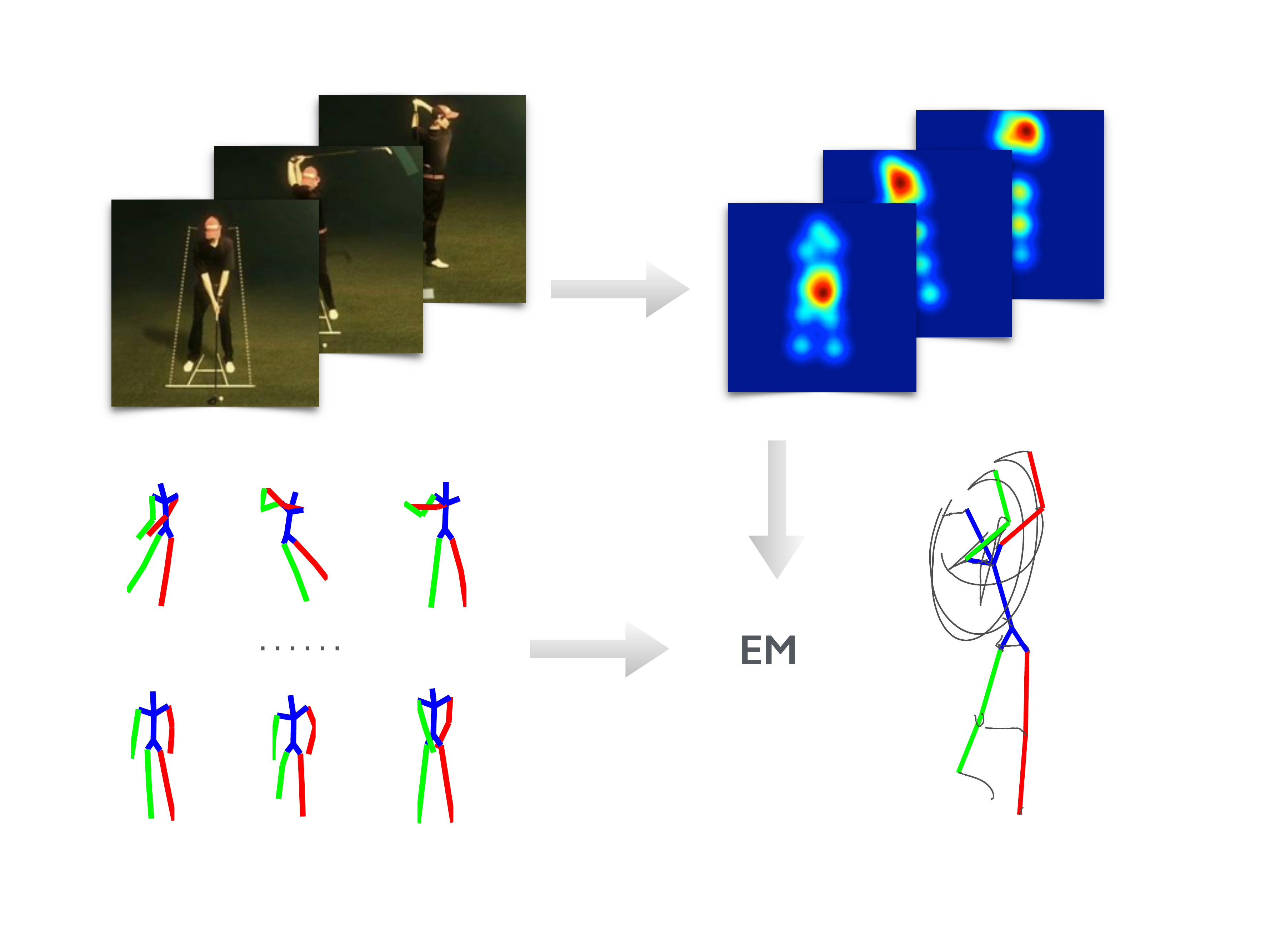}
   \caption{Overview of the proposed approach. (top-left) Input image sequence, (top-right) CNN-based  heat map outputs
   representing the soft localization of 2D joints,
   (bottom-left) 3D pose dictionary, and (bottom-right) the recovered 3D pose sequence reconstruction.}
   \label{fig:overview}
   \vspace{-1em}
\end{figure}

This paper is concerned with the challenge of recovering the 3D full-body human pose from a monocular RGB image sequence.
Potential applications of the presented research include human-computer interaction (cf.\ \cite{kinect}), surveillance, video browsing and indexing, and virtual reality.


From a geometric perspective, 3D articulated pose recovery is inherently ambiguous from monocular imagery \cite{lee1985determination}.
Further difficulties are raised due to the large variation in human appearance (e.g., clothing, body shape, and illumination), arbitrary camera viewpoint, and obstructed visibility due to external entities and self-occlusions. Notable successes in pose estimation consider the challenge of 2D pose recovery using discriminatively trained 2D part models coupled with 2D deformation priors, e.g., \cite{yang2011articulated,andriluka20142d,nie2015joint}, and more recently using deep learning, e.g., \cite{toshev2014deep}. Here, the 3D pose geometry is not leveraged. Combining robust image-driven 2D part detectors, expressive 3D geometric pose priors and temporal models to aggregate information over time is a promising area of research that has been given limited attention, e.g., \cite{andriluka2010monocular,zhou2014spatio}.  The challenge posed is how to seamlessly integrate 2D, 3D and temporal information to fully account for the model and measurement uncertainties.

This paper presents a 3D pose recovery framework that consists of a novel synthesis between
discriminative image-based and 3D reconstruction approaches.
In particular, the approach
reasons jointly about image-based 2D part location estimates and model-based 3D pose reconstruction, so that they can benefit from each other.  Further,
to improve the approach's robustness against detector error, occlusion, and reconstruction ambiguity, temporal smoothness is imposed on the 3D pose and viewpoint parameters.
\refFig{fig:overview} provides an overview of the proposed approach.
Given the input video (Fig.\ \ref{fig:overview}, top-left), 2D joint heat maps are generated with a deep convolutional neural network (CNN)  (Fig.\ \ref{fig:overview}, top-right).  These heat maps are combined with a sparse model of 3D human pose (Fig.\ \ref{fig:overview}, bottom-left) within an Expectation-Maximization (EM) framework to recover the 3D pose sequence (Fig.\ \ref{fig:overview}, bottom-right).



Considerable research has addressed the challenge of human motion capture from imagery \cite{moeslund2006survey,sminchisescu20073d,brubaker2010video,deva2011_Book}.
This work includes 2D human pose recovery in both single images (e.g., \cite{yang2011articulated,toshev2014deep,chen2014articulated,jain2014learning,tompson2014joint}) and video, e.g., \cite{sapp2011,cherian2014,nie2015joint,park2015articulated,pfister2015flowing,zhang2015human}.
In the current work, focus is placed on 3D pose recovery in video, where the pose model and prior are expressed in their natural 3D domain.

Early research on 3D monocular pose estimation in videos largely centred on incremental frame-to-frame pose tracking, e.g., \cite{bregler1998tracking,sminchisescu2003kinematic,sigal2012loose}.  These approaches rely on a given pose and dynamic model to constrain the pose search space.
Notable drawbacks of this approach include: the requirement that the initialization be provided and their inability to recover from tracking failures.  To address these limitations, more recent approaches have cast the tracking
problem as one of data association across frames, i.e., ``tracking-by-detection'', e.g., \cite{andriluka2010monocular}.  Here, candidate poses are first detected in each frame and subsequently a linking process
attempts to establish temporally consistent poses.

Another strand of research has focused on methods that predict 3D poses by searching a database of exemplars \cite{shakhnarovich2003fast,mori2006recovering,jiang20103d} or via a discriminatively learned mapping from the image directly or image features to human joint locations \cite{agarwal2006recovering,salzmann2010implicitly,yu2013unconstrained,ionescu2014human,tekin2015predicting}.
Recently, deep convolutional networks (CNNs) have emerged as a common element behind many state-of-the-art approaches, including human pose estimation, e.g., \cite{toshev2014deep,li20143d,tompson2014joint,li2015maximum}.  Here, two general approaches can be distinguished.  The first approach casts the pose estimation task as a joint location regression problem from the input image \cite{toshev2014deep,li20143d,li2015maximum}.  The second approach uses a CNN architecture for body part detection \cite{chen2014articulated,jain2014learning,tompson2014joint,pfister2015flowing}
and then typically enforces the 2D spatial relationship between body parts as a subsequent processing step.
Similar to the latter approaches, the proposed approach uses a CNN-based architecture to regress confidence heat maps of 2D joint position predictions.
The current work departs from these approaches by enforcing 3D spatial part relationships rather than 2D ones.

Most closely related to the present paper are generic factorization approaches for recovering 3D non-rigid shapes from image sequences captured with a single camera \cite{bregler2000recovering,akhter2011trajectory,dai2012simple,zhu2014complex,cho2015complex}, i.e., non-rigid structure from motion (NRSFM),
and human pose recovery models based on known skeletons \cite{lee1985determination,taylor2000reconstruction,valmadre2010deterministic,park20113d,leonardos2016articulated} or sparse representations \cite{ramakrishna2012reconstructing,fan2014pose,akhter2015pose,zhou20153d,zhou2015sparse}. Much of this work has been realized by assuming manually labeled 2D joint locations; however, there is some recent work that has used a 2D pose detector to automatically provide the input joints \cite{simo2012single,wang2014robust} or solved 2D and 3D pose estimation jointly \cite{simo2013joint,zhou2014spatio}.



\noindent{\bf Contributions:} The proposed approach advances the state-of-the-art in the following three ways.
First, in contrast to prediction methods (e.g., \cite{ionescu2014human,li2015maximum}), the proposed approach does not require synchronized 2D-3D data, as captured by motion capture systems. The proposed approach only requires readily available annotated 2D imagery (e.g., the ``in-the-wild'' PennAction dataset \cite{zhang2013actemes}) to train a CNN part detector and a separate 3D motion capture dataset (e.g., the CMU MoCap database) for the pose dictionary.
Second, in comparison to other 3D reconstruction methods (e.g., \cite{ramakrishna2012reconstructing,akhter2015pose}), the proposed approach considers
an arbitrary pose uncertainty. 
Finally, in contrast to prior work that consider two disjoint steps (i.e., detection of 2D joints and subsequent lifting the detections to 3D),  the current approach
combines these steps by  casting the 2D joint locations as latent variables.
This allows us to leverage the 3D geometric prior to help
2D joint localization and to rigorously handle the 2D estimation
uncertainty in a statistical framework.

\section{Models}
In this section, the models that describe the relationships between 3D poses, 2D poses and images are introduced.

\subsection{Sparse representation of 3D poses}

The 3D human pose is represented by the 3D locations of a set of $p$ joints, which is denoted by $\bfS_t\in\RR{3}{p}$ for frame $t$.
To reduce the ambiguity for 3D reconstruction, it is assumed that a 3D pose can be represented as a linear combination of predefined basis poses:
\begin{align}\label{eq:shape-model}
    \bfS_t = \sum_{i=1}^{k} c_{it}\bfB_i,
\end{align}
where $\bfB_i\in\RR{3}{p}$ denotes a basis pose and $c_{it}$ the corresponding weight. The basis poses are learned from training poses provided by a motion capture (MoCap) dataset.
Instead of using the conventional active shape model \cite{cootes1995}, where the basis set is small, a sparse representation is adopted which has proven in recent work to be
capable of modelling the large variability of human pose, e.g., \cite{ramakrishna2012reconstructing,akhter2015pose,zhou20153d}.
That is, an overcomplete dictionary, $\{\bfB_1,\cdots,\bfB_k\}$, is learned with a relatively large number of basis poses, $k$, where
the coefficients, $c_{it}$, are assumed to be sparse. In the remainder of this paper,
$\bfc_t$ denotes the coefficient vector $[c_{1t},\cdots,c_{kt}]^\top$ for frame $t$ and $\bfC$ denotes the matrix composed of all $\bfc_{t}$.

\subsection{Dependence between 2D and 3D poses}

The dependence between a 3D pose and its imaged 2D pose is modelled with a weak perspective camera model:
\begin{align}\label{eq:camera-model}
    \bfW_t = \bfR_t\bfS_t + \bfT_t\bfone^\top,
\end{align}
where $\bfW_t\in\RR{2}{p}$ denotes the 2D pose in frame $t$, and $\bfR_t\in\RR{2}{3}$ and $\bfT_t\in\R{2}$ the camera rotation and translation, respectively.
Note, the scale parameter in the weak perspective model is removed because the 3D structure, $\bfS_t$, can itself be scaled. In the following,
$\bfW$, $\bfR$ and $\bfT$ denote the collections of $\bfW_t$, $\bfR_t$ and $\bfT_t$ for all $t$, respectively.

Considering the observation noise and model error, the conditional distribution of the 2D poses given the 3D pose parameters is modelled as
\begin{align}\label{eq:likelihood}
\pr(\bfW|\theta) \propto e^{-\mathcal{L}(\theta;\bfW)},
\end{align}
where $\theta=\{\bfC,\bfR,\bfT\}$ is the union of all the 3D pose parameters
and the loss function, $\mathcal{L}(\theta;\bfW)$, is defined as
\begin{align}\label{eq:loss}
\mathcal{L}(\theta;\bfW) = \frac{\nu}{2}\sum_{t=1}^{n}\left\|\bfW_t - \bfR_t\sum_{i=1}^{k} c_{it}\bfB_i - \bfT_t\bfone^\top\right\|_F^2,
\end{align}
with $\|\cdot\|_F$ denoting the Frobenius norm. The model in \refEq{eq:likelihood} states that, given the 3D poses and camera parameters, the 2D location of each joint belongs to a Gaussian distribution with a mean equal to the projection of its 3D counterpart and a precision (i.e., the inverse variance) equal to $\nu$.

\subsection{Dependence between pose and image}

When 2D poses are given, it is assumed that the distribution of 3D pose parameters is conditionally independent of the image data. Therefore, the likelihood function of $\theta$ can be factorized as
\begin{align}\label{eq:likelihood-complete}
\pr(\bfI,\bfW|\theta) = \pr(\bfI|\bfW)\pr(\bfW|\theta),
\end{align}
where $\bfI=\{\bfI_1,\cdots,\bfI_n\}$ denotes the input images and $\pr(\bfW|\theta)$ is given in \refEq{eq:likelihood}. $\pr(\bfI|\bfW)$ is difficult to directly model, but it is proportional to $\pr(\bfW|\bfI)$ by assuming uniform priors on $\bfW$ and $\bfI$, and $\pr(\bfW|\bfI)$ can be learned from data.

Given the image data, the 2D distribution of each joint is assumed to be only dependent on the current image. Thus,
\begin{align}\label{eq:pr-I-W}
\pr(\bfI|\bfW) \propto \pr(\bfW|\bfI) = \Pi_{t}\Pi_{j}h_j(\bfw_{jt};\bfI_t),
\end{align}
where $\bfw_{jt}$ denotes the image location of joint $j$ in frame $t$, and $h_j(\cdot;\bfY)$ represents a mapping from an image $\bfY$ to a probability distribution of the joint location (termed heat map).
For each joint $j$, the mapping $h_j$ is approximated  by a CNN learned from training data. The details of CNN learning are described in \refSec{sec:cnn}.

\subsection{Prior on model parameters}

The following penalty function on the model parameters is introduced:
\begin{align}
\mathcal{R}(\theta) = \alpha\|\bfC\|_1 + \frac{\beta}{2}\|\nabla_t\bfC\|_F^2 + \frac{\gamma}{2}\|\nabla_t\bfR\|_F^2,
\end{align}
where $\|\cdot\|_1$ denotes the $\ell_1$-norm (i.e., the sum of absolute values), and $\nabla_t$ the discrete temporal derivative operator.
The first term penalizes the cardinality of the pose coefficients to induce a sparse pose representation. The second and third terms impose first-order smoothness on both the pose coefficients and rotations.

\section{3D pose inference}
In this section, the proposed approach to 3D pose inference is described.  Here, two cases are distinguished: (i) the image locations of the joints are provided (Section\ \ref{sec:known}) and
(ii) the joint locations are unknown (Section \ref{sec:unknown}).

\subsection{Given 2D poses}\label{sec:known}

When the 2D poses, $\bfW$, are given, the model parameters, $\theta$, are recovered via penalized maximum likelihood estimation (MLE):
\begin{align}\label{eq:pmle}
\theta^* &= \argmax_{\theta} ~ \ln\pr(\bfW|\theta) - \mathcal{R}(\theta) \nonumber \\
& = \argmin_{\theta} ~ \mathcal{L}(\theta;\bfW) + \mathcal{R}(\theta).
\end{align}
The problem in \refEq{eq:pmle} is solved via block coordinate descent, i.e., alternately updating $\bfC$, $\bfR$ or $\bfT$ while fixing the others. The update of $\bfC$ needs to solve:
\begin{align}\label{eq:C-step}
\bfC \leftarrow \argmin_{\bfC} ~~ \mathcal{L}(\bfC;\bfW) + \alpha\|\bfC\|_1 + \frac{\beta}{2}\|\nabla_t\bfC\|_F^2,
\end{align}
where the objective is the composite of two differentiable functions plus an $\ell_1$ penalty. The problem in \refEq{eq:C-step} is solved by accelerated proximal gradient (APG) \cite{nesterov2007gradient}.
Since the problem in \refEq{eq:C-step}  is convex, global optimality is guaranteed. The update of $\bfR$ needs to solve:
\begin{align}\label{eq:R-step}
\bfR \leftarrow \argmin_{\bfR} ~~ \mathcal{L}(\bfR;\bfW) + \frac{\gamma}{2}\|\nabla_t\bfR\|_F^2,
\end{align}
where the objective is differentiable and the variables are rotations restricted to $SO(3)$. Here, manifold optimization is adopted to update the rotations
using the trust-region solver in the Manopt toolbox \cite{boumal2014manopt}.
The update of $\bfT$ has the following closed-form solution:
\begin{align}\label{eq:T-step}
\bfT_t \leftarrow \mbox{row mean}\left\{\bfW_t - \bfR_t\sum_{i=1}^{k} c_{it}\bfB_i\right\}.
\end{align}

The entire algorithm for 3D pose inference given the 2D poses is summarized in \refAlg{alg:bcd}.
The iterations are terminated once the objective value has converged.
Since in each step the objective function is non-increasing, the algorithm is guaranteed to converge; however, since the problem in \refEq{eq:pmle} is nonconvex, the algorithm requires
a suitably chosen initialization (described in \refSec{sec:initialization}).

\begin{algorithm}[t]\label{alg:bcd}
\LinesNumbered
\caption{Block coordinate descent to solve \refEq{eq:pmle}.}
\vspace{0.3em}
\KwIn{$\bfW$ \tcp*[r]{\small 2D joint locations}}
\KwOut{$\bfC,\bfR,\bfT$ \tcp*[r]{\small pose parameters}}
\vspace{0.3em}
initialize the parameters \tcp*{\small \refSec{sec:initialization}}
\While{not converged}{
update $\bfC$ by \refEq{eq:C-step} with APG\;
update $\bfR$ by \refEq{eq:R-step} with Manopt\;
update $\bfT$ by \refEq{eq:T-step};
}
\end{algorithm}

\subsection{Unknown 2D poses}\label{sec:unknown}

If the 2D poses are unknown, $\bfW$ is treated as a latent variable and is marginalized during the estimation process. The marginalized likelihood function is
\begin{align}
\pr(\bfI|\theta) = \int\pr(\bfI,\bfW|\theta)d\bfW, \label{eq:marginalization}
\end{align}
where $\pr(\bfI,\bfW|\theta)$ is given in \refEq{eq:likelihood-complete}.

Direct marginalization of (\ref{eq:marginalization}) is extremely difficult.
Instead, an EM algorithm is developed to compute the penalized MLE.
In the expectation step, the expectation of the penalized log-likelihood is calculated with respect to the conditional distribution of $\bfW$ given the image data and the previous estimate of all the 3D pose parameters, $\theta'$:
\begin{align}
&Q(\theta|\theta') = \int \left\{\ln\pr(\bfI,\bfW|\theta) - \mathcal{R}(\theta)\right\} ~ \pr(\bfW|\bfI,\theta') d\bfW \nonumber \\
&= \int \left\{\ln\pr(\bfI|\bfW) + \ln\pr(\bfW|\theta) - \mathcal{R}(\theta)\right\} \pr(\bfW|\bfI,\theta') d\bfW \nonumber \\
&= \mbox{const} - \int \mathcal{L}(\theta;\bfW) \pr(\bfW|\bfI,\theta') d\bfW - \mathcal{R}(\theta).
\end{align}

It can be easily shown that
\begin{align}\label{eq:der1}
\int \mathcal{L}(\theta;\bfW) \pr(\bfW|\bfI,\theta') d\bfW = \mathcal{L}(\theta;\mathbb{E}\left[\bfW|\bfI,\theta'\right]) + \mbox{const},
\end{align}
where $\mathbb{E}\left[\bfW|\bfI,\theta'\right]$ is the expectation of $\bfW$ given $\bfI$ and $\theta'$:
\begin{align}\label{eq:der2}
\mathbb{E}\left[\bfW|\bfI,\theta'\right]
&= \int \pr(\bfW|\bfI,\theta')~ \bfW ~ d\bfW \nonumber \\
&= \int \frac{\pr(\bfI|\bfW)\pr(\bfW|\theta')}{Z} ~ \bfW ~ d\bfW,
\end{align}
and $Z$ is a scalar that normalizes the probability. The derivation of \refEq{eq:der1} and \refEq{eq:der2} is given in the supplementary material.
Both $\pr(\bfI|\bfW)$ and $\pr(\bfW|\theta')$ given in \refEq{eq:pr-I-W} and \refEq{eq:likelihood}, respectively, are products of marginal probabilities of $\bfw_{jt}$. Therefore,
the expectation of each $\bfw_{jt}$ can be computed separately. In particular, the expectation of each $\bfw_{jt}$ is efficiently approximated by sampling over the pixel grid.

In the maximization step, the following is computed:
\begin{align}
\theta \leftarrow &\argmax_{\theta} Q(\theta|\theta') \nonumber \\
= &\argmin_{\theta} ~~ \mathcal{L}(\theta;\mathbb{E}\left[\bfW|\bfI,\theta'\right]) + \mathcal{R}(\theta),
\end{align}
which can be solved by \refAlg{alg:bcd}.

The entire EM algorithm is summarized in \refAlg{alg:em} with the initialization scheme described next in \refSec{sec:initialization}.

\begin{algorithm}[t]\label{alg:em}
\LinesNumbered
\caption{The EM algorithm for pose from video.}
\vspace{0.3em}
\KwIn{$h_j(\cdot;\bfI_t),~ \forall j,t$ \tcp*[r]{\small heat maps}}
\KwOut{$\theta=\{\bfC,\bfR,\bfT\}$ \tcp*[r]{\small pose parameters}}
\vspace{0.3em}
initialize the parameters \tcp*[r]{\small \refSec{sec:initialization}}
\While{not converged}{
$\theta'=\theta$\;
\vspace{0.3em}
\tcp{\small Compute the expectation of $\bfW$}
$\mathbb{E}\left[\bfW|\bfI,\theta'\right] = \int \frac{1}{Z}\pr(\bfI|\bfW)\pr(\bfW|\theta') ~ \bfW ~ d\bfW$\;
\vspace{0.3em}
\tcp{\small Update $\theta$ by \refAlg{alg:bcd}}
$\theta = \argmin_{\theta} ~~ \mathcal{L}(\theta;\mathbb{E}\left[\bfW|\bfI,\theta'\right]) + \mathcal{R}(\theta)$ \;
}
\vspace{0.5em}
\end{algorithm}

\subsection{Initialization}\label{sec:initialization}

A convex relaxation approach \cite{zhou20153d,zhou2015sparse} is used to initialize the parameters. In \cite{zhou20153d},
a convex formulation was proposed to solve the single frame pose estimation problem given 2D correspondences, which is a special case of \refEq{eq:pmle}.
The approach was later extended to handle 2D correspondence outliers \cite{zhou2015sparse}. If the 2D poses are given, the model parameters are initialized for each frame separately with the convex method proposed in \cite{zhou20153d}.
Alternatively, if the 2D poses are unknown, for each joint, the image location with the maximum heat map value is used.
Next, the robust estimation algorithm from \cite{zhou2015sparse} is applied to initialize the parameters.

\section{CNN-based joint uncertainty regression}\label{sec:cnn}
A CNN is used to learn the mapping $\bfY \mapsto h_j(\cdot;\bfY)$, where $\bfY$ denotes an input image and $h_j(\cdot;\bfY)$ represents a heat map for joint $j$. Instead of learning $p$ networks for $p$ joints, a fully convolutional neural network \cite{long2015fully} is trained to regress $p$ joint distributions simultaneously by taking into account the full-body information.

During training, a rectangular patch is extracted around the subject from each image and is resized to $256 \times 256$  pixels. Random shifts are applied during cropping and RGB channel-wise random noise is added for data augmentation. Channel-wise RGB mean values are computed from the dataset and subtracted from the images for data normalization. The training labels to be regressed are multi-channel heat maps with each channel corresponding to the image location uncertainty distribution for each joint. The uncertainty is modelled by a Gaussian centered at the annotated joint location with variance $\sigma = 1.5$. The heat map resolution is reduced to $32 \times 32$ to decrease the CNN model size which allows a large batch size in training and prevents overfitting.

The CNN architecture used is similar to the SpatialNet model proposed elsewhere \cite{pfister2015flowing} but without any spatial fusion or temporal pooling. The network consists of seven convolutional layers with $5 \times 5$ filters followed by ReLU layers and a last convolutional layer with $1\times1\times p$ filters to provide dense prediction for all joints. A $2\times2$ max pooling layer is inserted after each of the first three convolutional layers. The network is trained by minimizing the $l_2$ loss between the prediction and the label with the open source Caffe framework \cite{jia2014caffe}. Stochastic gradient descent (SGD) with momentum of 0.9 and a mini-batch size of 128 is used. 

During testing, consistent with previous 3D pose methods (e.g., \cite{li2015maximum,tekin2015predicting}), a bounding box around the subject is assumed and the image patch in the bounding box $\bfI_t$ is cropped in frame $t$ and fed forward through the network to predict the heat maps, $h_j(\cdot;\bfI_t),~\forall j=1,\dotsc,n$.

\section{Empirical evaluation}
\subsection{Datasets and implementation details}

Empirical evaluation was performed on two datasets -- Human3.6M \cite{ionescu2014human} and PennAction \cite{zhang2013actemes}.

The Human3.6M dataset \cite{ionescu2014human} is a recently published large-scale dataset for 3D human sensing. It includes millions of 3D human poses acquired from a MoCap system with corresponding images from calibrated cameras.
This setup provides synchronized videos and 2D-3D pose data for evaluation. It includes 11 subjects performing 15 actions, 
such as eating, sitting and walking. 
The same data partition protocol as in previous work was used \cite{li2015maximum,tekin2015predicting}: the data from five subjects (S1, S5, S6, S7, S8) was used for training and the data from two subjects (S9, S11) was used for testing. The original frame rate is 50 fps and is downsampled to 10 fps.

The PennAction dataset \cite{zhang2013actemes} is a recently introduced in-the-wild human action dataset
containing 2326 challenging consumer videos.  The dataset consists of
15 actions, such as golf swing, bowling, and tennis swing. Each of the video sequences is manually annotated frame-by-frame with 13 human body joints in 2D.
In evaluation, PennAction's training and testing split was used which consists of an even split of the videos between training and testing.

The algorithm in \cite{zhou2015sparse} was used to learn the pose dictionaries. The dictionary size was set to $K=64$ for action-specific dictionaries and $K=128$ for the nonspecific action case.
For all experiments, the parameters of the proposed model were fixed ($\alpha=0.1$, $\beta=5$, $\gamma=0.5$, $\nu=4$ in a normalized 2D coordinate system).

\subsection{Evaluation with known 2D poses} 

First, the evaluation of the 3D reconstructability of the proposed method with known 2D poses is presented. The generic approach to 3D reconstruction from 2D correspondences across a sequence is NRSFM. The proposed method is compared to the state-of-the-art method for NRSFM \cite{dai2012simple} on the Human3.6M dataset. A recent baseline method for single-view pose reconstruction Projected Matching Pursuit (PMP) \cite{ramakrishna2012reconstructing} is also included in comparison.

The sequences of S9 and S11 from the first camera in the Human 3.6M dataset were used for evaluation and frames beyond 30 seconds were truncated for each sequence. The 2D orthographic projections of the 3D poses provided in the dataset were used as the input. Performance was evaluated by the mean per joint error (mm) in 3D by comparing the reconstructed pose against the ground truth. As the standard protocol for evaluating NRSFM, the error was calculated up to a similarity transformation via the Procrustes analysis. To demonstrate the generality of the proposed approach, a single pose dictionary from all the training pose data, irrespective of the action type, was used, i.e., a non-action specific model. The method from Dai et al.\ \cite{dai2012simple} requires a predefined rank $K$.  Here, various
values of $K$ were considered with the best result for each sequence reported.


\begin{table}
\centering
\renewcommand{\arraystretch}{1.2}
\begin{tabular}{l*{15}{c}}
\toprule
 & Original & Synthesized \\
\toprule
PMP \cite{ramakrishna2012reconstructing} & 89.50 & 84.16 \\
NRSFM \cite{dai2012simple} & 72.98 & 48.88 \\
Single frame initialization & 50.04 & 48.08 \\
Optimization by \refAlg{alg:bcd} & \textbf{49.64} & \textbf{47.57} \\
\toprule
\end{tabular}
\vspace{0.25em}
\caption{3D reconstruction given 2D poses. Two input cases are considered: original 2D pose data from Human3.6M and synthesized 2D pose data with artificial camera motion. The numbers are the mean per joint errors (mm) in 3D. }\label{tab:nrsfm}
\vspace{-1em}
\end{table}

\begin{table*}
\renewcommand{\arraystretch}{1.2}
\begin{tabular}{l*{15}{c}}
\toprule
 & Directions & Discussion & Eating & Greeting & Phoning & Photo & Posing & Purchases \\
\toprule
LinKDE \cite{ionescu2014human} & 132.71 & 183.55 & 132.37 & 164.39 & 162.12 & 205.94 & 150.61 & 171.31 \\
Li et al. \cite{li2015maximum} & - & 136.88 & 96.94 & 124.74 & - & 168.68 & - & - \\
Tekin et al. \cite{tekin2015predicting} & 102.39 & 158.52 & 87.95 & 126.83 & 118.37 & 185.02 & 114.69 & 107.61 \\
Proposed & \textbf{87.36} & \textbf{109.31} & \textbf{87.05} & \textbf{103.16} & \textbf{116.18} &  \textbf{143.32} & \textbf{106.88} & \textbf{99.78} \\
\toprule
 & Sitting & SittingDown & Smoking & Waiting & WalkDog & Walking & WalkTogether & Average \\
\toprule
LinKDE \cite{ionescu2014human} & 151.57 & 243.03 & 162.14 & 170.69 & 177.13 & 96.60 & 127.88 & 162.14  \\
Li et al. \cite{li2015maximum} & - & - & - & - & 132.17 & 69.97 & - & - \\
Tekin et al. \cite{tekin2015predicting} & 136.15 & 205.65 & 118.21 & 146.66 & 128.11 & \textbf{65.86} & \textbf{77.21} & 125.28  \\
Proposed & \textbf{124.52} & \textbf{199.23} & \textbf{107.42} & \textbf{118.09} & \textbf{114.23} & 79.39 & 97.70 & \textbf{113.01} \\
\toprule
\end{tabular}
\vspace{0.25em}
\caption{Quantitative comparison on Human 3.6M datasets. The numbers are the mean per joint errors (mm) in 3D evaluated for different actions of Subjects 9 and 11. }\label{tab:h36m}
\vspace{-1em}
\end{table*}

The results are shown in the second column of \refTab{tab:nrsfm}. The proposed method clearly outperforms the NRSFM baseline. The reason is that the videos are captured by stationary cameras. Although the subject is occasionally rotating, the ``baseline" between frames is generally small, and neighboring views provide insufficient geometric constraints for 3D reconstruction. In other words, NRSFM is very difficult to compute with slow camera motion. This observation is consistent with prior findings in the NRSFM literature, e.g., \cite{akhter2011trajectory}. To validate this issue, an artificial rotation was applied to the 3D poses by 15 degrees per second and the 2D joint locations were synthesized by projecting the rotated 3D poses into 2D. The corresponding results are presented in the third column of \refTab{tab:nrsfm}. In this case, the performance of NRSFM improved dramatically. Overall, the experiments demonstrate that the structure prior (even a non-action specific one) from existing pose data is critical for reconstruction.
This is especially true for videos with small camera motion, which is common in real world applications. The temporal smoothness helps but the change is not significant since the single frame initialization is very stable with known 2D poses. Nevertheless,
in the next section it is shown that the temporal smoothness is important when 2D poses are not given.

\subsection{Evaluation with unknown poses: Human3.6M }\label{sec:h36m}

\begin{table}
\centering
\renewcommand{\arraystretch}{1.2}
\begin{tabular}{l*{15}{c}}
\toprule
 & 3D (mm) & 2D (pixel) \\
\toprule
Single frame initialization & 143.85 & 15.00 \\
Optimization by \refAlg{alg:em} & 125.55 & 10.85 \\
Perspective adjustment & \textbf{113.01} & \textbf{10.85} \\
\hline
No smoothness & 120.99 & 11.25 \\
No action label & 116.49 & 10.87 \\
\toprule
\end{tabular}
\vspace{0.25em}
\caption{The estimation errors after separate steps and under additional settings. The numbers are the average per joint errors for all testing data in both 3D and 2D.}\label{tab:steps}
\vspace{-1em}
\end{table}

\begin{figure*}
  \centering
  \includegraphics[width=0.42\linewidth]{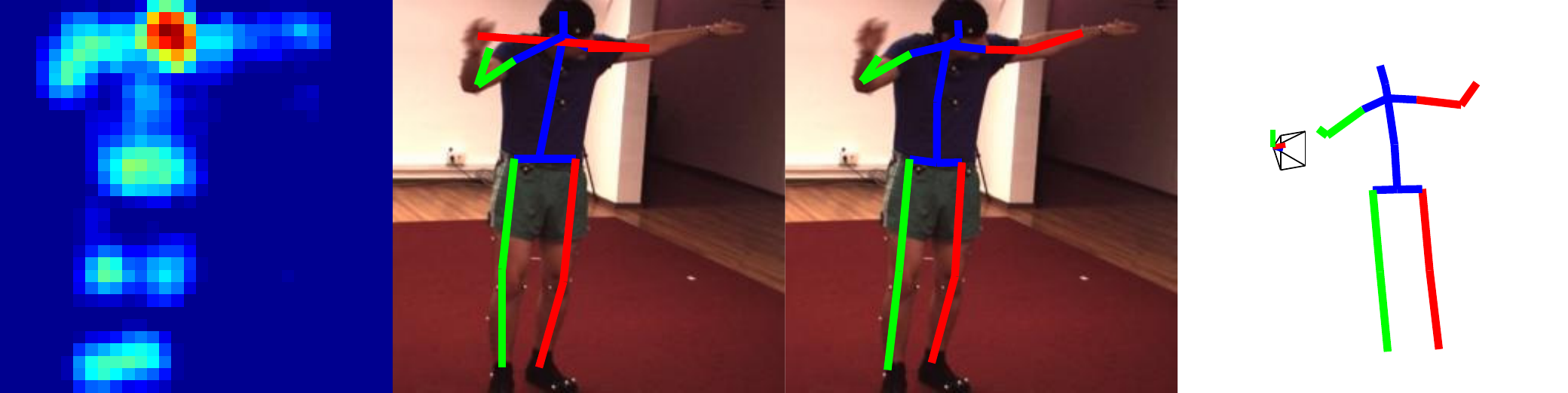}\hspace{3em}
  \includegraphics[width=0.42\linewidth]{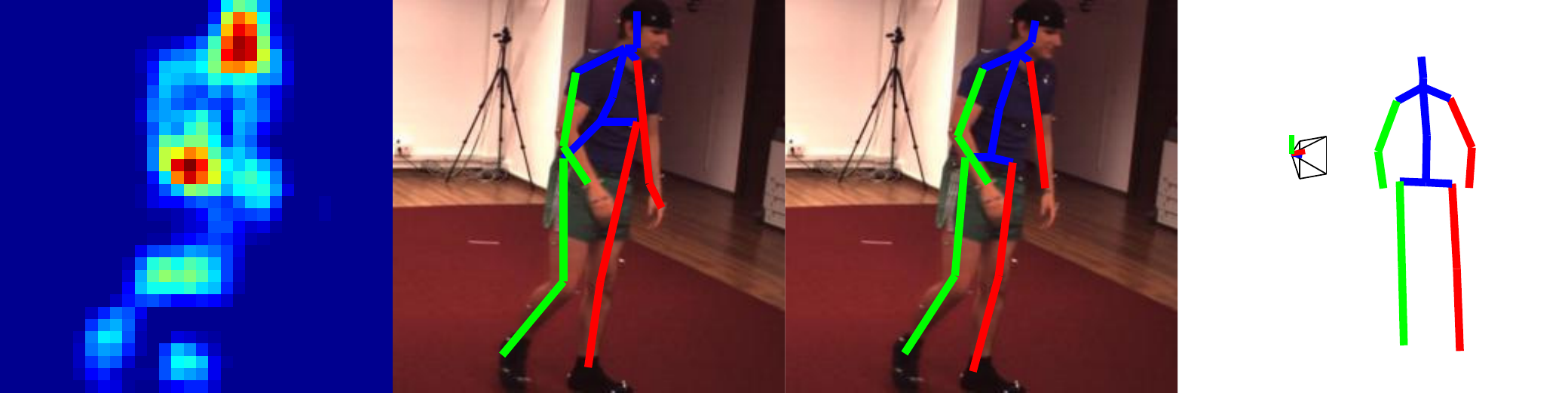}
  \includegraphics[width=0.42\linewidth]{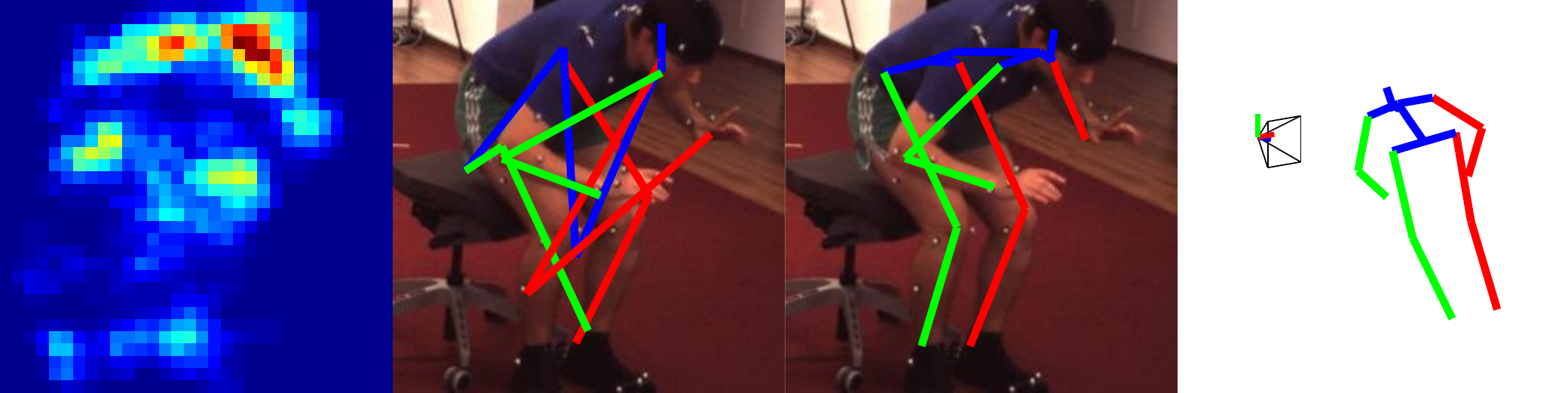}\hspace{3em}
  \includegraphics[width=0.42\linewidth]{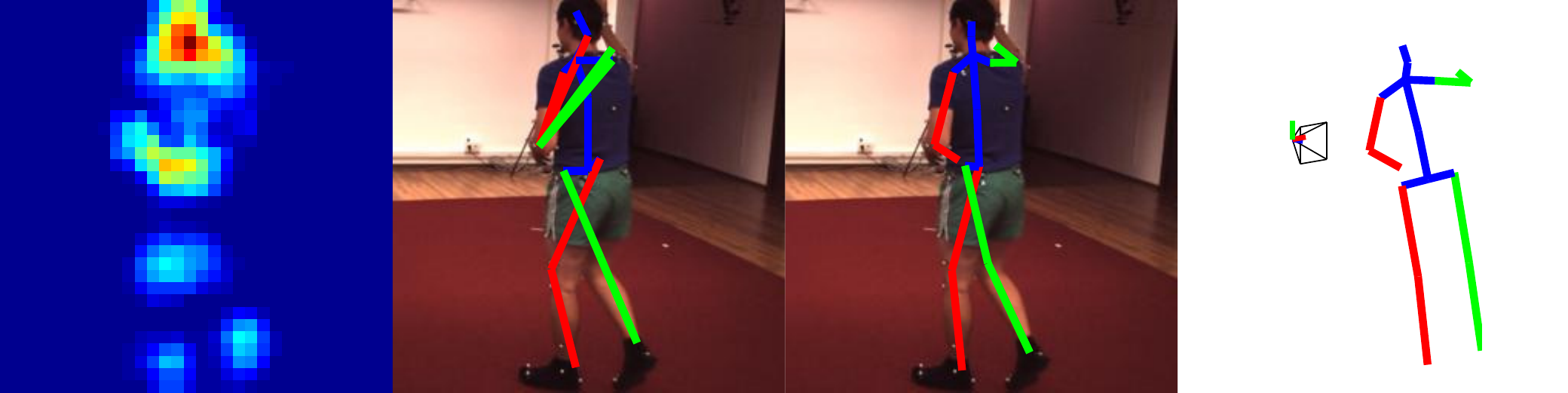}
  \includegraphics[width=0.42\linewidth]{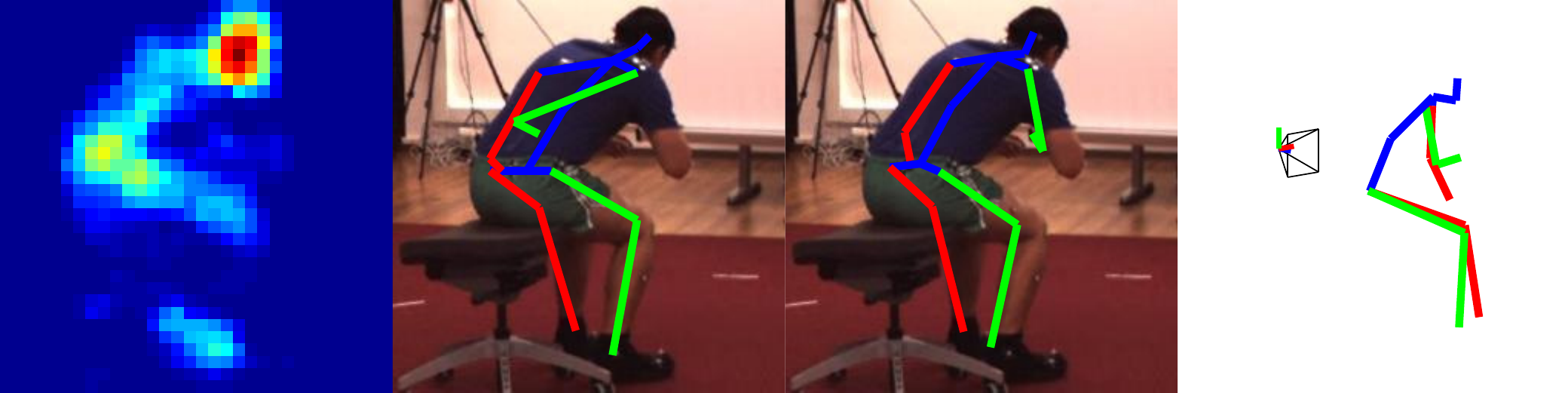}\hspace{3em}
  \includegraphics[width=0.42\linewidth]{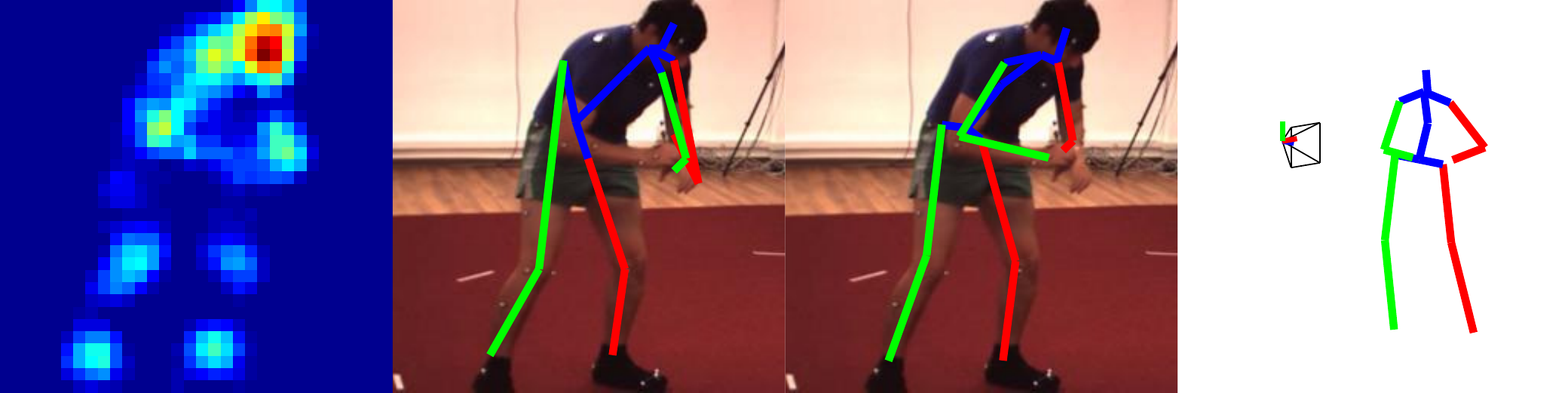}
  \includegraphics[width=0.42\linewidth]{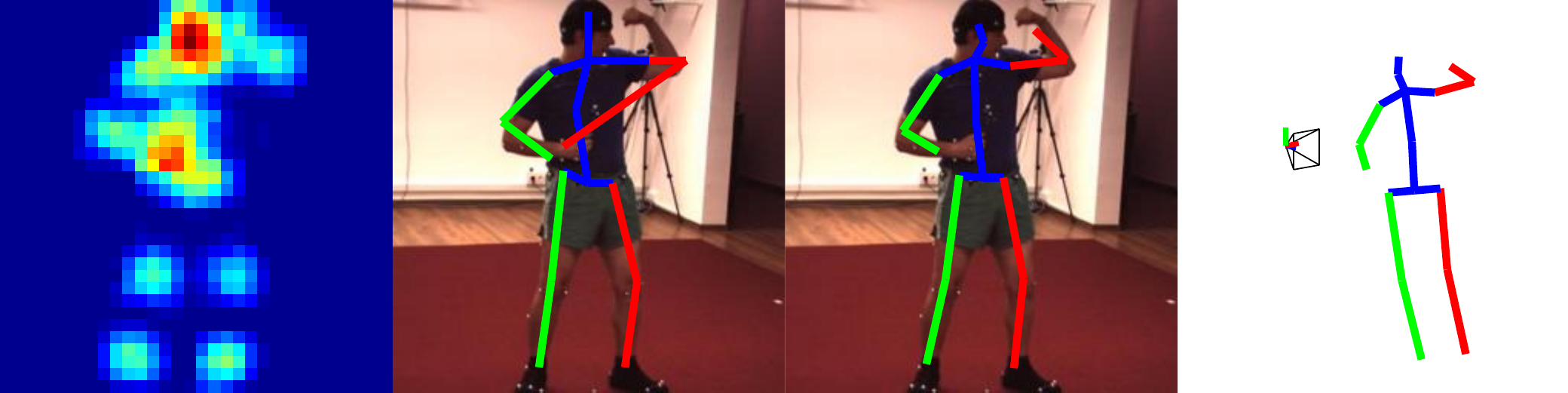}\hspace{3em}
  \includegraphics[width=0.42\linewidth]{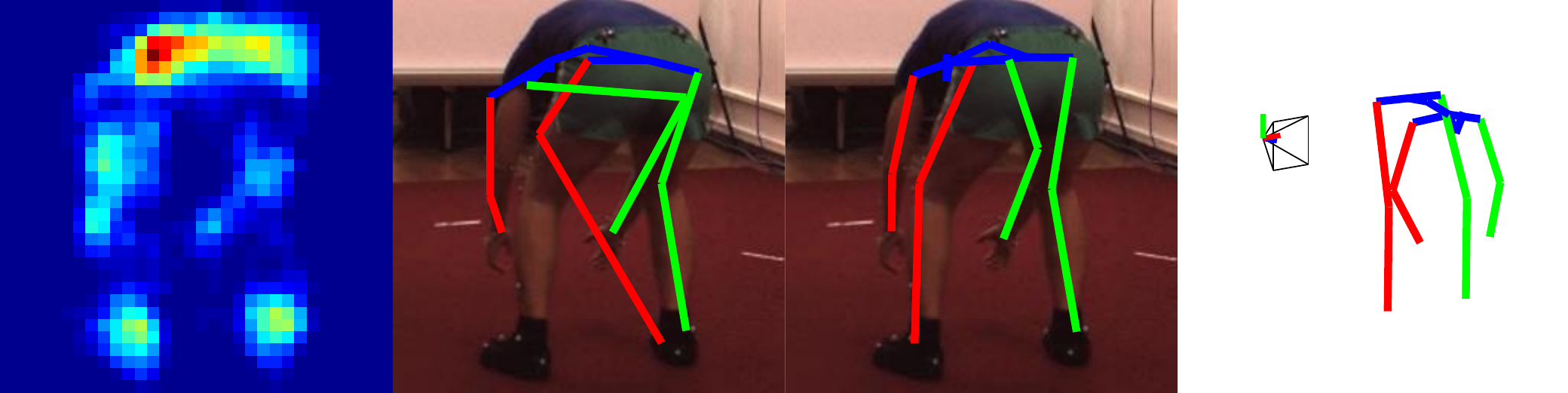}
  \includegraphics[width=0.42\linewidth]{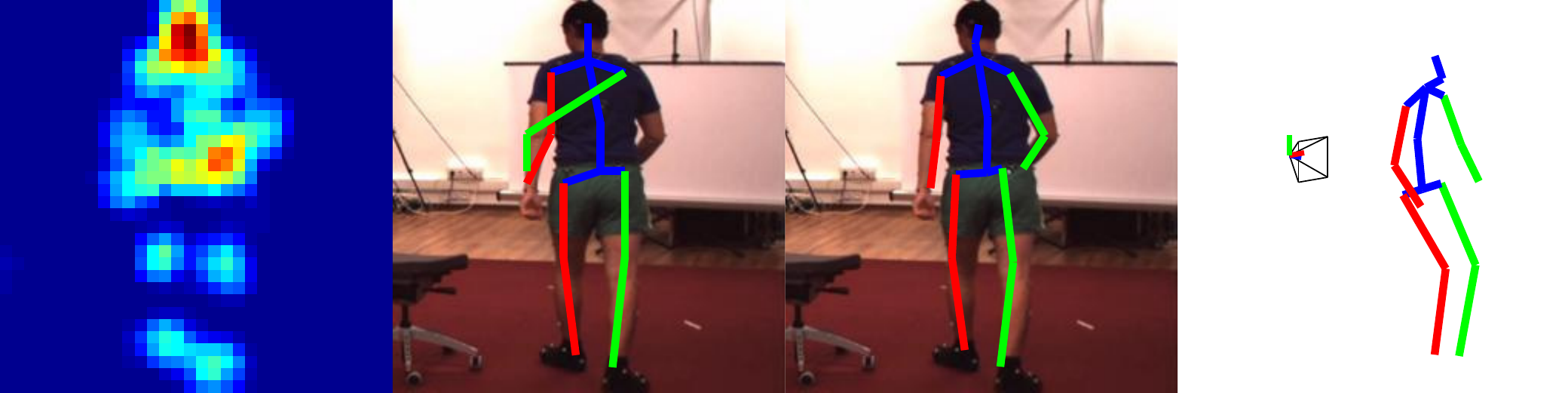}\hspace{3em}
  \includegraphics[width=0.42\linewidth]{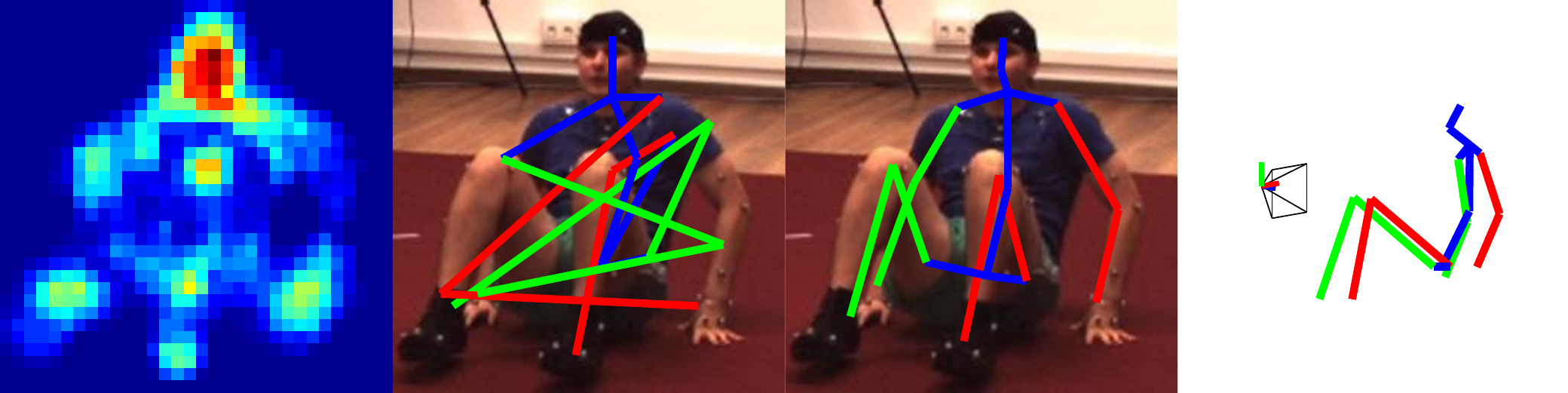}
  \includegraphics[width=0.42\linewidth]{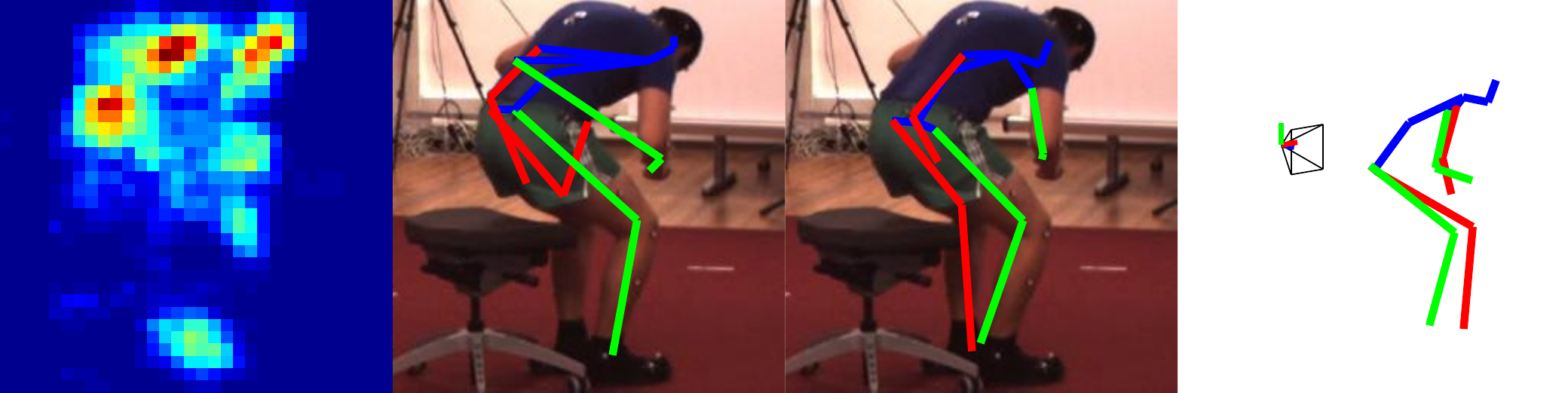}\hspace{3em}
  \includegraphics[width=0.42\linewidth]{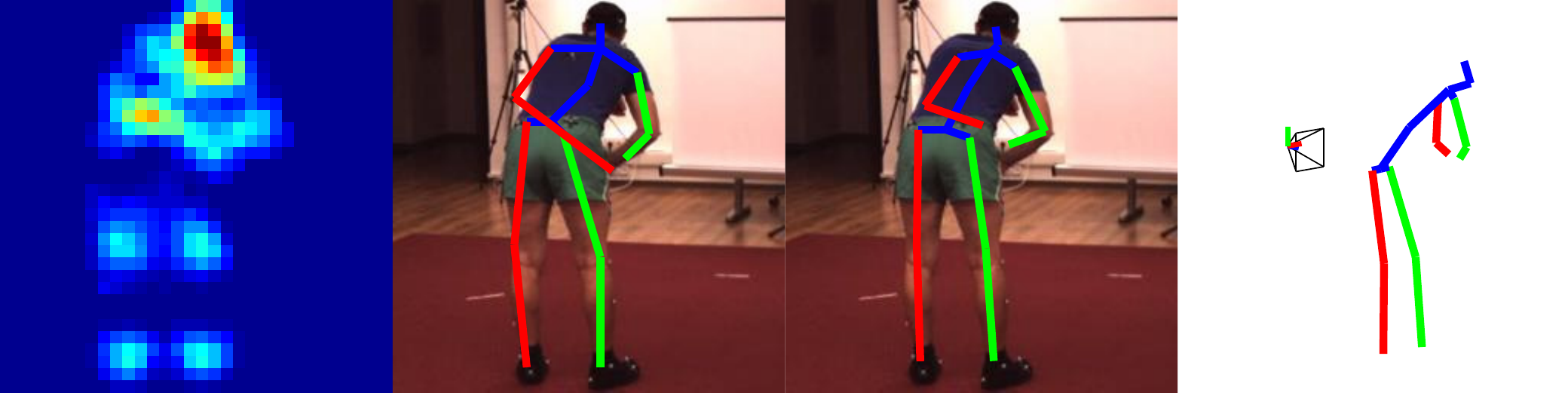}
  \includegraphics[width=0.42\linewidth]{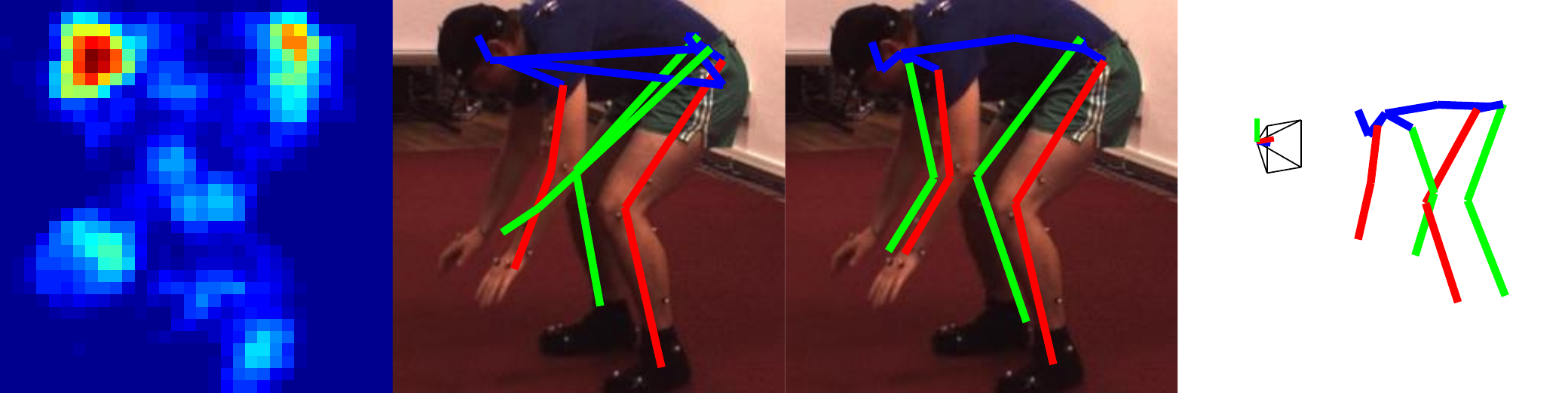}\hspace{3em}
  \includegraphics[width=0.42\linewidth]{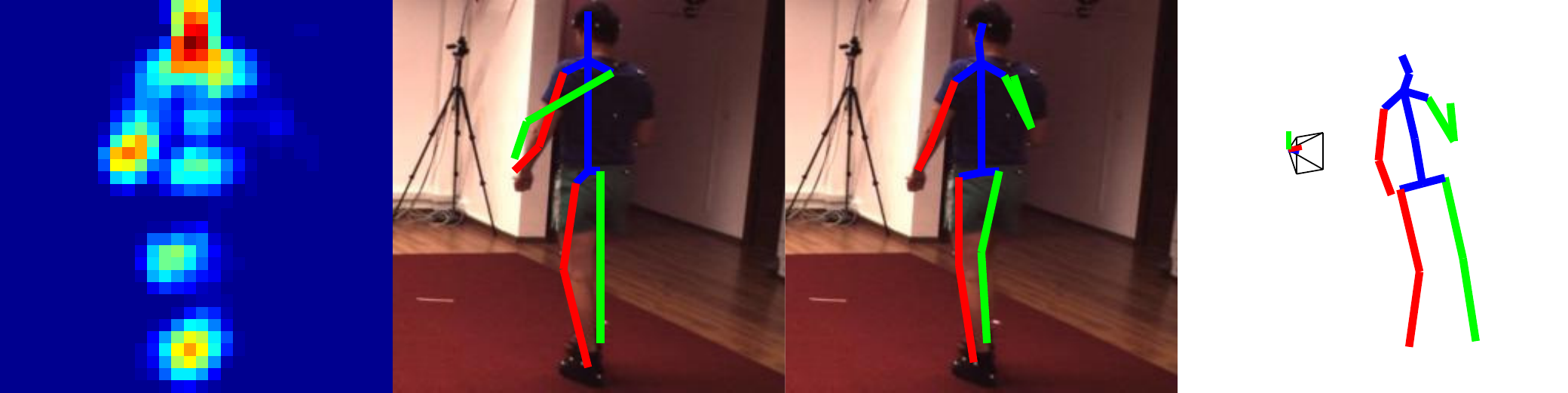}
  \caption{Example frame results on Human3.6M, where the errors in the 2D heat maps are corrected after considering the pose and temporal smoothness priors. Each row includes two examples from two actions. 
 The figures from left-to-right correspond to the heat map (all joints combined), the 2D pose by greedily locating each joint separately according to the heat map, the estimated 2D pose by the proposed EM algorithm, and the estimated 3D pose visualized in a novel view. The original viewpoint is also shown. }\label{fig:h36m}
\end{figure*}

Next, results on the Human3.6M dataset are reported when 2D poses are not given. The proposed method is compared to three recent baseline methods. The first baseline method is LinKDE which is provided with the Human3.6M dataset \cite{ionescu2014human}.  This baseline is based on single frame regression. The second one is from Tekin et al.\ \cite{tekin2015predicting} which extends the first baseline method by exploring motion information in a short sequence. The third one is a recently published CNN-based method from Li et al.\ \cite{li2015maximum}.

In this experiment, the sequences of S9 and S11 from all cameras were used for evaluation. The standard evaluation protocol of the Human3.6M dataset was adopted, i.e., the mean per joint error (mm) in 3D is calculated between the reconstructed pose and the ground truth in the camera frame with their root locations aligned. Note that the Procrustes alignment is not allowed here. In general, it is impossible to determine the scale of the object in monocular images. The baseline methods learned the scale from training subjects.  For a fair comparison, the reconstructed pose by the proposed method was scaled such that the mean limb length of the reconstructed pose was identical to the average value of all training subjects. As the alignment to the ground truth was not allowed, the joint error was largely affected by the camera rotation estimate, and empirically the misalignment was largely due to the adopted weak perspective camera model. To compensate the misalignment, the rotation estimate was refined for each frame with a perspective camera model (the 2D and 3D human pose estimates were fixed) by a perspective-n-point (PnP) algorithm \cite{lu2000fast}

The results are summarized in \refTab{tab:h36m}. The table shows that the proposed method achieves the best results on most of the actions except for ``walk" and ``walk together", which involve very predictable and repetitive motions and might favor the direct regression approach \cite{tekin2015predicting}. In addition, the results of the proposed approach have the smallest variation across all actions with a standard deviation of  $28.75$ versus $37.80$ from Tekin et al.

In \refTab{tab:steps},  3D reconstruction and 2D joint localization results are provided under several setup variations of the proposed approach.
Note that the 2D errors are with respect to the normalized bounding box size $256\times256$. The table shows that the convex initialization provides suitable initial estimates, which are further improved by the EM algorithm that integrates joint detection uncertainty and temporal smoothness. The perspective adjustment is important under the Human3.6M evaluation protocol, where Procrustes alignment to the ground truth is not allowed.  The proposed approach was also evaluated under two additional settings.
In the first setting, the smoothness constraint was removed from the proposed model by setting $\beta=\gamma=0$.  As a result, the average error significantly increased.
This demonstrates the importance of incorporating temporal smoothness. In the second setting, a single CNN and pose dictionary was learned from all training data.  These models were then applied
to all testing data without distinguishing the videos by their action class.
As a result, the estimation error increased, which is attributed to the fact that the 3D reconstruction ambiguity is greatly enlarged if the pose prior is not restricted to an action class.

\refFig{fig:h36m} visualizes the results of some example frames. While the heat maps may be erroneous due to occlusion, left-right ambiguity, and other uncertainty from the detectors, the proposed EM algorithm can largely correct the errors by leveraging the pose prior, integrating temporal smoothness, and modelling the uncertainty.

\subsection{Evaluation with unknown poses: PennAction}

\begin{figure*}
  \centering
  \includegraphics[width=0.45\linewidth]{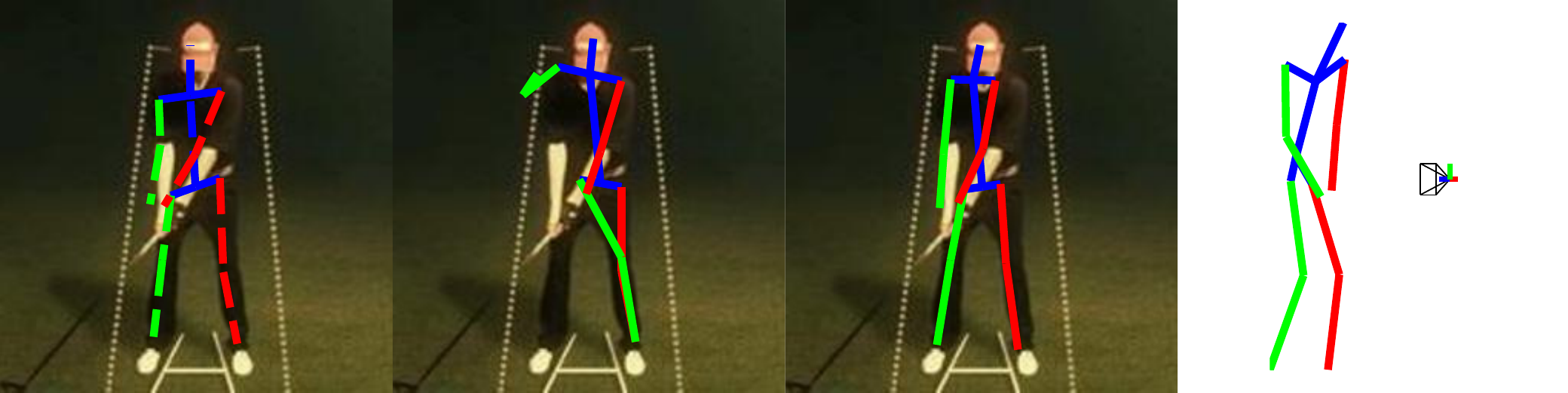}\hspace{2em}
  \includegraphics[width=0.45\linewidth]{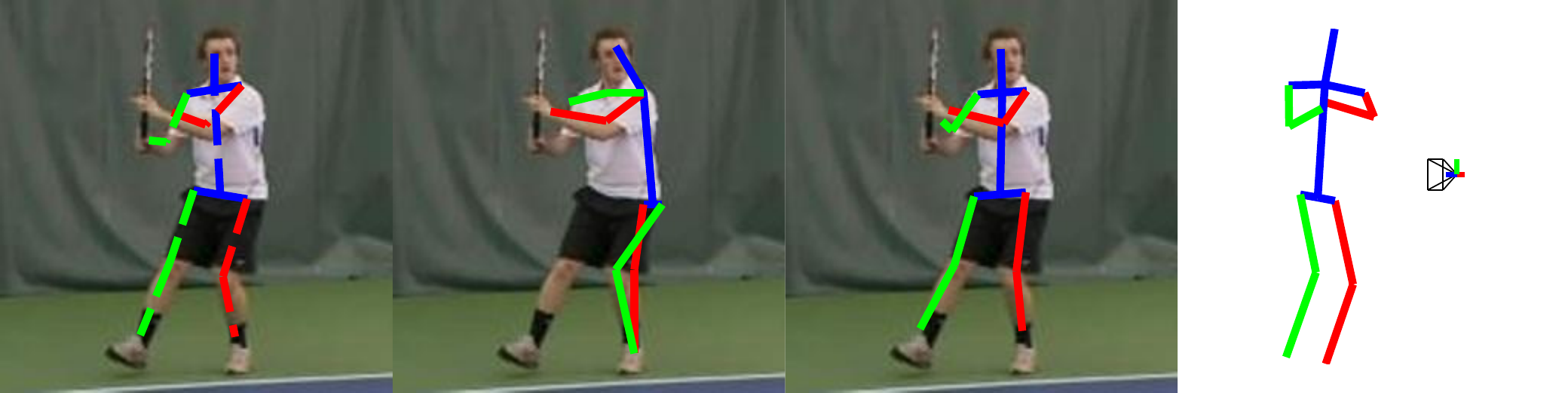}
  \includegraphics[width=0.45\linewidth]{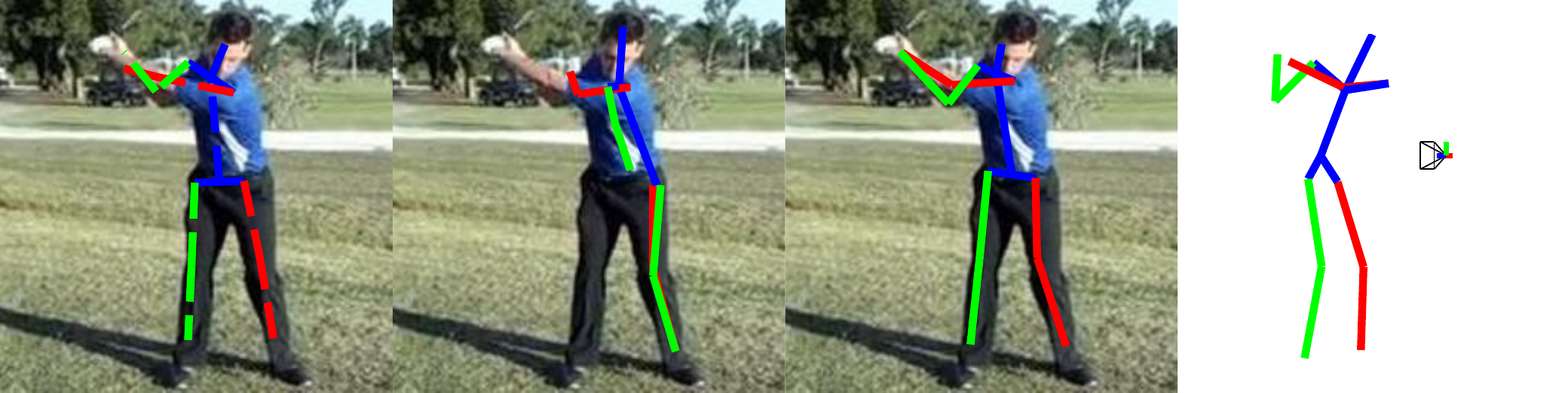}\hspace{2em}
  \includegraphics[width=0.45\linewidth]{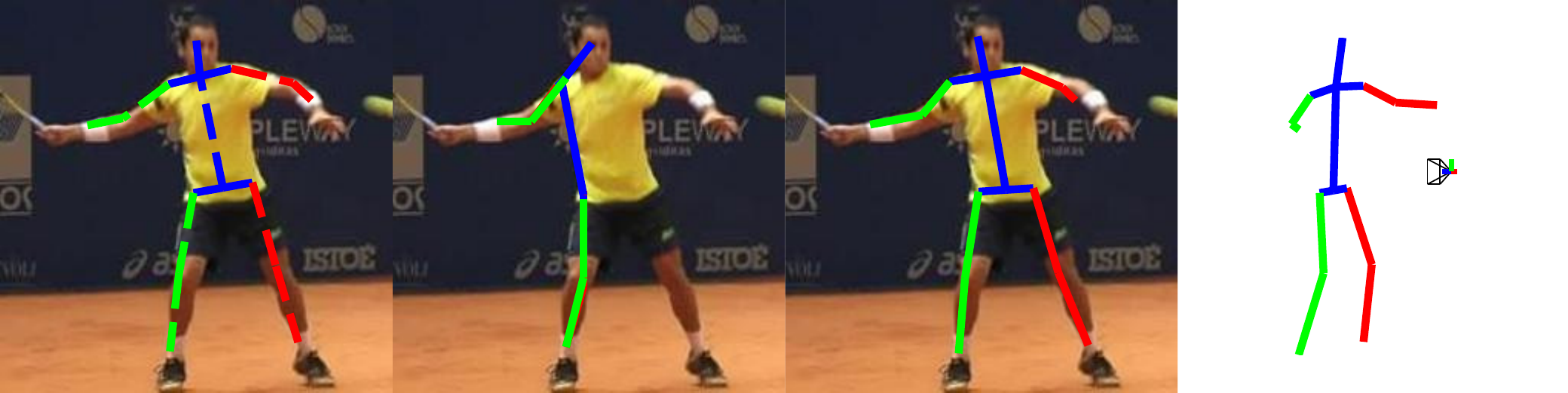}
  \includegraphics[width=0.45\linewidth]{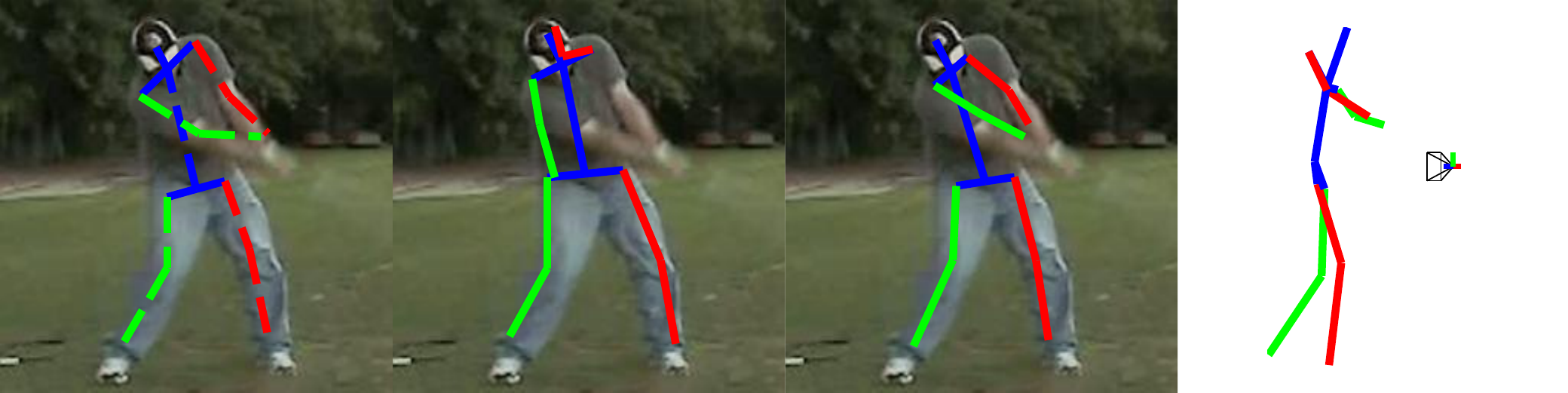}\hspace{2em}
  \includegraphics[width=0.45\linewidth]{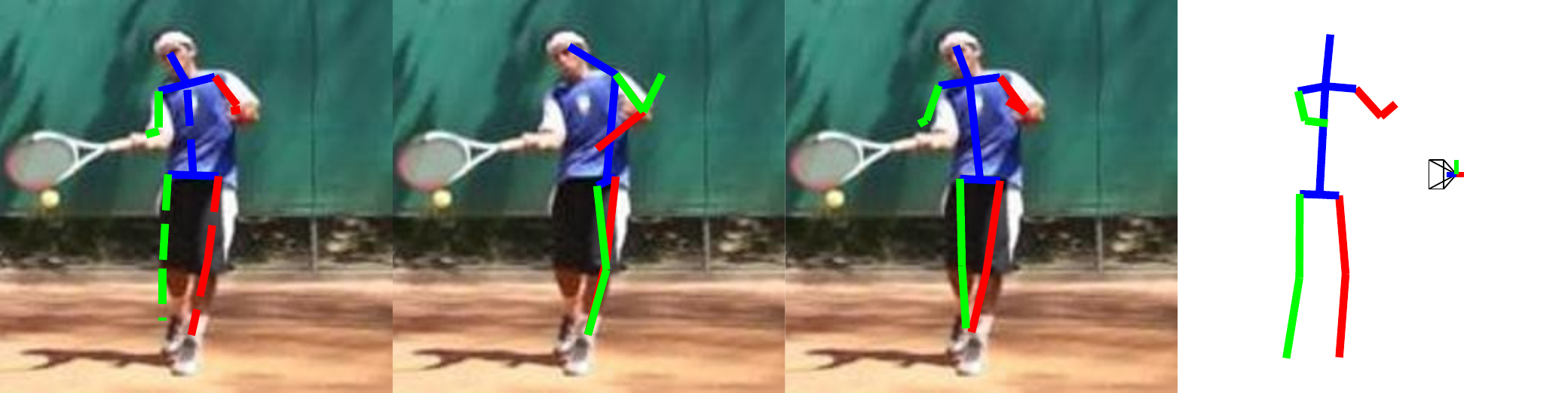}
  \includegraphics[width=0.45\linewidth]{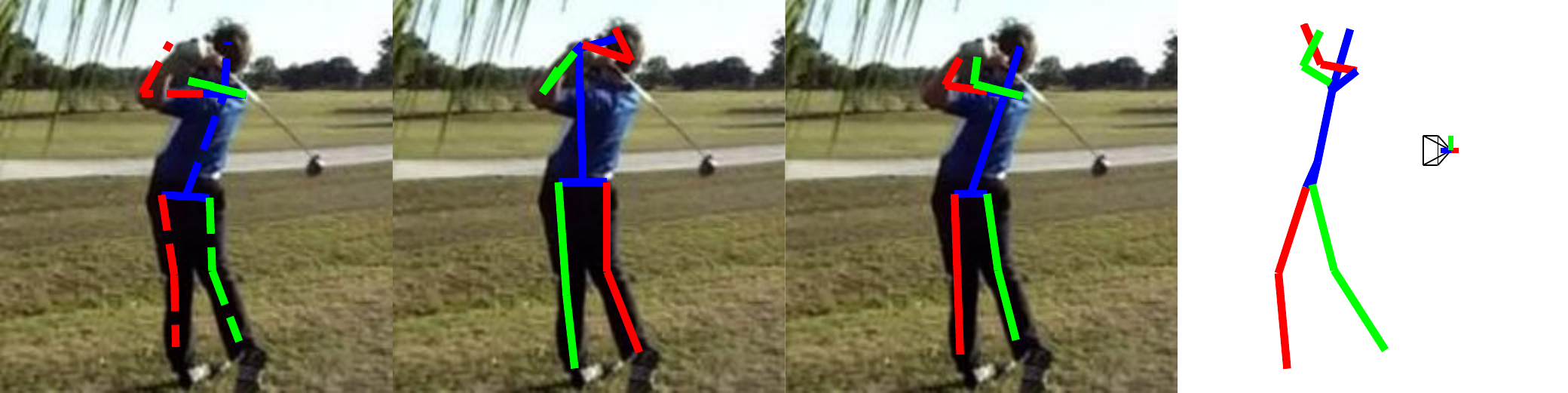}\hspace{2em}
  \includegraphics[width=0.45\linewidth]{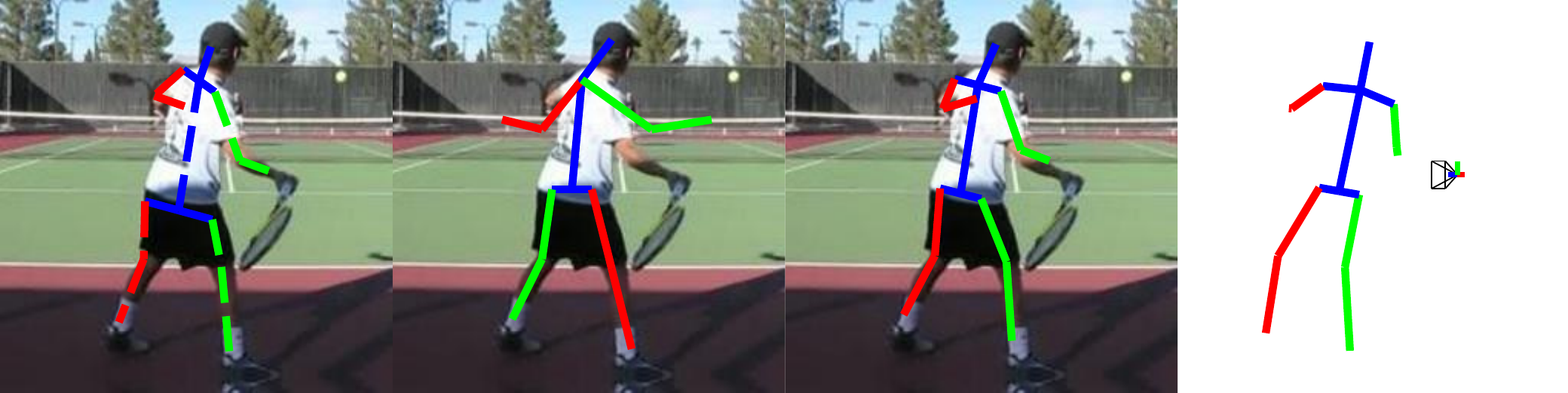}
  \caption{Example results on PennAction. Each row includes two examples. In each example, the figures from left-to-right correspond to the ground truth superimposed on the image, the estimated pose using the baseline approach \cite{yang2011articulated}, the estimated pose by the proposed approach, and the estimated 3D pose visualized in a novel view. The original viewpoint is also shown. }\label{fig:penn}
\end{figure*}

Finally, the applicability of the proposed approach for pose estimation with in-the-wild videos is demonstrated. Results are reported using two actions from the PennAction dataset:
``golf swing" and ``tennis forehand", both of which are very challenging due to large pose variability, self-occlusion, and image blur caused by fast motion.
For the proposed approach, the CNN was trained using the annotated training images from the PennAction dataset, while the pose dictionary was learned with publicly available MoCap data\footnote{Data sources: \url{http://mocap.cs.cmu.edu} and \url{http://www.motioncapturedata.com}}. Due to the lack of 3D ground truth, quantitative 2D pose estimation results are reported and compared with the publicly available 2D pose detector from Yang and Ramanan \cite{yang2011articulated}.  The baseline was retrained on the PennAction dataset. Note that the baseline methods considered in \refSec{sec:h36m} are not applicable here since they require synchronized 2D image and 3D pose data for training.

To measure joint localization accuracy, both the widely used per joint distance errors and the probability of correct keypoint (PCK) metrics are used.
The PCK metric measures the fraction of correctly located joints with respect to a threshold. Here, the threshold is set to $10$ pixels which is roughly the half length of a head segment.

 \refTab{tab:penn} summarizes the quantitative results.
The initialization step alone outperformed the baseline. This demonstrates the effectiveness of CNN-based approaches, which has been shown in many recent works, e.g., \cite{toshev2014deep,pfister2015flowing}. The proposed EM algorithm further improves upon the initialization results by a large margin by integrating the geometric and smoothness priors.
Several example results are shown in \refFig{fig:penn}. It can be seen that the proposed method successfully recovers the poses for various subjects under a variety of viewpoints. In particular, compared to the baseline, the proposed method does not suffer from the well-known ``double-counting'' problem for tree-based models \cite{yang2011articulated} due to the holistic 3D pose prior.

\begin{table}
\centering
\renewcommand{\arraystretch}{1.2}
\begin{tabular}{l*{15}{c}}
\toprule
 & Baseline & Initial & Optimized \\
\toprule
Golf & 24.78 / 0.38 & 18.73 / 0.45 & \textbf{14.03 / 0.54} \\
Tennis & 29.15 / 0.40 & 25.75 / 0.42 & \textbf{20.99 / 0.45} \\
\toprule
\end{tabular}
\vspace{0.25em}
\caption{2D pose errors on PennAction. Each pair of numbers correspond to the per joint distance error (pixels) and the PCK metric. The baseline is the retrained model from Yang and  Ramanan \cite{yang2011articulated}. The last two columns correspond to the errors after initialization and EM optimization in the proposed approach.}\label{tab:penn}
\vspace{-1em}
\end{table}

\subsection{Running time}

The experiments were performed on a desktop with an Intel i7 3.4G CPU, 8G RAM and a TitanZ GPU.
The running times for CNN-based heat map generation and convex initialization were roughly 1s and 0.6s per frame, respectively; both steps can be easily parallelized. The EM algorithm usually converged in 20 iterations with a CPU time less than 100s for a sequence of 300 frames.

\section{Summary}
In summary, a  3D pose estimation framework from video has been presented that consists of a novel synthesis between a deep learning-based 2D part
regressor, a sparsity-driven 3D reconstruction approach and a 3D temporal smoothness prior.  This joint consideration combines the discriminative power of state-of-the-art 2D part detectors,
the expressiveness of 3D pose models and regularization by way of aggregating information over time.
In practice, alternative joint detectors, pose representations and temporal models can be conveniently integrated in the proposed framework by replacing the original components.  Experiments demonstrated that 3D geometric priors and temporal coherence can not only help 3D reconstruction but also improve 2D joint localization. Future extensions may include incremental algorithms for online tracking-by-detection and handling multiple subjects.  

\small

\bigskip
\noindent\textbf{Supplementary material}:
The MATLAB code, evaluation on the HumanEva I dataset, demonstration videos, and other supplementary materials are available at: \url{http://cis.upenn.edu/~xiaowz/monocap.html}.

\bigskip
\noindent\textbf{Acknowledgments}: The authors are grateful for support through the following grants:
NSF-DGE-0966142, NSF-IIS-1317788, NSF-IIP-1439681, NSF-IIS-1426840, ARL MAST-CTA W911NF-08-2-0004, ARL RCTA W911NF-10-2-0016, ONR N000141310778, and NSERC Discovery.

\newpage
\small
\bibliographystyle{ieee}
\bibliography{bibref_definitions_short,bibref}

\begin{thebibliography}{10}\itemsep=-1pt

\bibitem{agarwal2006recovering}
A.~Agarwal and B.~Triggs.
\newblock Recovering 3{D} human pose from monocular images.
\newblock {\em PAMI}, 28(1):44--58, 2006.

\bibitem{akhter2015pose}
I.~Akhter and M.~J. Black.
\newblock Pose-conditioned joint angle limits for 3{D} human pose
  reconstruction.
\newblock In {\em CVPR}, 2015.

\bibitem{akhter2011trajectory}
I.~Akhter, Y.~Sheikh, S.~Khan, and T.~Kanade.
\newblock Trajectory space: {A} dual representation for nonrigid structure from
  motion.
\newblock {\em PAMI}, 33(7):1442--1456, 2011.

\bibitem{andriluka20142d}
M.~Andriluka, L.~Pishchulin, P.~Gehler, and B.~Schiele.
\newblock {2D} human pose estimation: New benchmark and state of the art
  analysis.
\newblock In {\em CVPR}, 2014.

\bibitem{andriluka2010monocular}
M.~Andriluka, S.~Roth, and B.~Schiele.
\newblock Monocular 3{D} pose estimation and tracking by detection.
\newblock In {\em CVPR}, 2010.

\bibitem{boumal2014manopt}
N.~Boumal, B.~Mishra, P.-A. Absil, and R.~Sepulchre.
\newblock Manopt, a {M}atlab toolbox for optimization on manifolds.
\newblock {\em JMLR}, 15:1455--1459, 2014.

\bibitem{bregler2000recovering}
C.~Bregler, A.~Hertzmann, and H.~Biermann.
\newblock Recovering non-rigid 3{D} shape from image streams.
\newblock In {\em CVPR}, 2000.

\bibitem{bregler1998tracking}
C.~Bregler and J.~Malik.
\newblock Tracking people with twists and exponential maps.
\newblock In {\em CVPR}, 1998.

\bibitem{brubaker2010video}
M.~A. Brubaker, L.~Sigal, and D.~J. Fleet.
\newblock Video-based people tracking.
\newblock In {\em Handbook of Ambient Intelligence and Smart Environments},
  pages 57--87. Springer, 2010.

\bibitem{chen2014articulated}
X.~Chen and A.~Yuille.
\newblock Articulated pose estimation by a graphical model with image dependent
  pairwise relations.
\newblock In {\em NIPS}, 2014.

\bibitem{cherian2014}
A.~Cherian, J.~Mairal, K.~Alahari, and C.~Schmid.
\newblock Mixing body-part sequences for human pose estimation.
\newblock In {\em CVPR}, pages 2361--2368, 2014.

\bibitem{cho2015complex}
J.~Cho, M.~Lee, and S.~Oh.
\newblock Complex non-rigid {3D} shape recovery using a {P}rocrustean normal
  distribution mixture model.
\newblock {\em IJCV}, pages 1--21, 2015.

\bibitem{cootes1995}
T.~F. Cootes, C.~J. Taylor, D.~H. Cooper, and J.~Graham.
\newblock Active shape models--{T}heir training and application.
\newblock {\em CVIU}, 61(1):38--59, 1995.

\bibitem{dai2012simple}
Y.~Dai, H.~Li, and M.~He.
\newblock A simple prior-free method for non-rigid structure-from-motion
  factorization.
\newblock {\em IJCV}, 107(2):101--122, 2014.

\bibitem{fan2014pose}
X.~Fan, K.~Zheng, Y.~Zhou, and S.~Wang.
\newblock Pose locality constrained representation for {3D} human pose
  reconstruction.
\newblock In {\em ECCV}, 2014.

\bibitem{ionescu2014human}
C.~Ionescu, D.~Papava, V.~Olaru, and C.~Sminchisescu.
\newblock Human3.6m: Large scale datasets and predictive methods for 3{D} human
  sensing in natural environments.
\newblock {\em PAMI}, 36(7):1325--1339, 2014.

\bibitem{jain2014learning}
A.~Jain, J.~Tompson, M.~Andriluka, G.~Taylor, and C.~Bregler.
\newblock Learning human pose estimation features with convolutional networks.
\newblock In {\em ICLR}, 2014.

\bibitem{jia2014caffe}
Y.~Jia, E.~Shelhamer, J.~Donahue, S.~Karayev, J.~Long, R.~Girshick,
  S.~Guadarrama, and T.~Darrell.
\newblock Caffe: Convolutional architecture for fast feature embedding.
\newblock {\em arXiv preprint arXiv:1408.5093}, 2014.

\bibitem{jiang20103d}
H.~Jiang.
\newblock 3{D} human pose reconstruction using millions of exemplars.
\newblock In {\em ICPR}, 2010.

\bibitem{lee1985determination}
H.~Lee and Z.~Chen.
\newblock Determination of {3D} human body postures from a single view.
\newblock {\em CVGIP}, 30(2):148--168, 1985.

\bibitem{leonardos2016articulated}
S.~Leonardos, X.~Zhou, and K.~Daniilidis.
\newblock Articulated motion estimation from a monocular image sequence using
  spherical tangent bundles.
\newblock In {\em ICRA}, 2016.

\bibitem{li20143d}
S.~Li and A.~B. Chan.
\newblock {3D} human pose estimation from monocular images with deep
  convolutional neural network.
\newblock In {\em ACCV}, 2014.

\bibitem{li2015maximum}
S.~Li, W.~Zhang, and A.~B. Chan.
\newblock Maximum-margin structured learning with deep networks for 3{D} human
  pose estimation.
\newblock In {\em ICCV}, 2015.

\bibitem{long2015fully}
J.~Long, E.~Shelhamer, and T.~Darrell.
\newblock Fully convolutional networks for semantic segmentation.
\newblock In {\em CVPR}, 2015.

\bibitem{lu2000fast}
C.-P. Lu, G.~D. Hager, and E.~Mjolsness.
\newblock Fast and globally convergent pose estimation from video images.
\newblock {\em PAMI}, 22(6):610--622, 2000.

\bibitem{moeslund2006survey}
T.~B. Moeslund, A.~Hilton, and V.~Kr\"{u}ger.
\newblock A survey of advances in vision-based human motion capture and
  analysis.
\newblock {\em CVIU}, 104(2):90--126, 2006.

\bibitem{mori2006recovering}
G.~Mori and J.~Malik.
\newblock Recovering 3{D} human body configurations using shape contexts.
\newblock {\em PAMI}, 28(7):1052--1062, 2006.

\bibitem{nesterov2007gradient}
Y.~Nesterov.
\newblock Gradient methods for minimizing composite objective function.
\newblock Technical report, Universit{\'e} catholique de Louvain, Center for
  Operations Research and Econometrics (CORE), 2007.

\bibitem{park2015articulated}
D.~Park and D.~Ramanan.
\newblock Articulated pose estimation with tiny synthetic videos.
\newblock In {\em ChaLearn Workshop on Looking at People, CVPR}, 2015.

\bibitem{park20113d}
H.~S. Park and Y.~Sheikh.
\newblock {3D} reconstruction of a smooth articulated trajectory from a
  monocular image sequence.
\newblock In {\em ICCV}, pages 201--208, 2011.

\bibitem{pfister2015flowing}
T.~Pfister, J.~Charles, and A.~Zisserman.
\newblock Flowing convnets for human pose estimation in videos.
\newblock In {\em ICCV}, 2015.

\bibitem{ramakrishna2012reconstructing}
V.~Ramakrishna, T.~Kanade, and Y.~Sheikh.
\newblock Reconstructing 3{D} human pose from 2{D} image landmarks.
\newblock In {\em ECCV}, 2012.

\bibitem{deva2011_Book}
D.~Ramanan.
\newblock Part-based models for finding people and estimating their pose.
\newblock In {\em Visual Analysis of Humans - Looking at People}, pages
  199--223. Springer, 2011.

\bibitem{salzmann2010implicitly}
M.~Salzmann and R.~Urtasun.
\newblock Implicitly constrained {G}aussian process regression for monocular
  non-rigid pose estimation.
\newblock In {\em NIPS}, 2010.

\bibitem{sapp2011}
B.~Sapp, D.~J. Weiss, and B.~Taskar.
\newblock Parsing human motion with stretchable models.
\newblock In {\em CVPR}, pages 1281--1288, 2011.

\bibitem{shakhnarovich2003fast}
G.~Shakhnarovich, P.~A. Viola, and T.~Darrell.
\newblock Fast pose estimation with parameter-sensitive hashing.
\newblock In {\em ICCV}, 2003.

\bibitem{kinect}
J.~Shotton, A.~W. Fitzgibbon, M.~Cook, T.~Sharp, M.~Finocchio, R.~Moore,
  A.~Kipman, and A.~Blake.
\newblock Real-time human pose recognition in parts from single depth images.
\newblock In {\em CVPR}, 2011.

\bibitem{sigal2012loose}
L.~Sigal, M.~Isard, H.~W. Haussecker, and M.~J. Black.
\newblock Loose-limbed people: Estimating 3{D} human pose and motion using
  non-parametric belief propagation.
\newblock {\em IJCV}, 98(1):15--48, 2012.

\bibitem{simo2013joint}
E.~Simo-Serra, A.~Quattoni, C.~Torras, and F.~Moreno-Noguer.
\newblock {A Joint Model for 2D and 3D Pose Estimation from a Single Image}.
\newblock In {\em CVPR}, 2013.

\bibitem{simo2012single}
E.~Simo-Serra, A.~Ramisa, G.~Aleny\`a, C.~Torras, and F.~Moreno-Noguer.
\newblock {Single Image 3D Human Pose Estimation from Noisy Observations}.
\newblock In {\em CVPR}, 2012.

\bibitem{sminchisescu20073d}
C.~Sminchisescu.
\newblock 3{D} human motion analysis in monocular video techniques and
  challenges.
\newblock In {\em AVSS}, 2007.

\bibitem{sminchisescu2003kinematic}
C.~Sminchisescu and B.~Triggs.
\newblock Kinematic jump processes for monocular 3{D} human tracking.
\newblock In {\em CVPR}, 2003.

\bibitem{taylor2000reconstruction}
C.~Taylor.
\newblock Reconstruction of articulated objects from point correspondences in a
  single uncalibrated image.
\newblock {\em CVIU}, 80(3):349--363, 2000.

\bibitem{tekin2015predicting}
B.~Tekin, X.~Sun, X.~Wang, V.~Lepetit, and P.~Fua.
\newblock Predicting people's 3{D} poses from short sequences.
\newblock {\em arXiv preprint arXiv:1504.08200}, 2015.

\bibitem{tompson2014joint}
J.~J. Tompson, A.~Jain, Y.~LeCun, and C.~Bregler.
\newblock Joint training of a convolutional network and a graphical model for
  human pose estimation.
\newblock In {\em NIPS}, 2014.

\bibitem{toshev2014deep}
A.~Toshev and C.~Szegedy.
\newblock Deep{P}ose: Human pose estimation via deep neural networks.
\newblock In {\em CVPR}, 2014.

\bibitem{valmadre2010deterministic}
J.~Valmadre and S.~Lucey.
\newblock Deterministic 3{D} human pose estimation using rigid structure.
\newblock In {\em ECCV}, 2010.

\bibitem{wang2014robust}
C.~Wang, Y.~Wang, Z.~Lin, A.~L. Yuille, and W.~Gao.
\newblock Robust estimation of 3{D} human poses from a single image.
\newblock In {\em CVPR}, 2014.

\bibitem{nie2015joint}
B.~Xiaohan~Nie, C.~Xiong, and S.-C. Zhu.
\newblock Joint action recognition and pose estimation from video.
\newblock In {\em CVPR}, 2015.

\bibitem{yang2011articulated}
Y.~Yang and D.~Ramanan.
\newblock Articulated pose estimation with flexible mixtures-of-parts.
\newblock In {\em CVPR}, 2011.

\bibitem{yu2013unconstrained}
T.~Yu, T.~Kim, and R.~Cipolla.
\newblock Unconstrained monocular 3{D} human pose estimation by action
  detection and cross-modality regression forest.
\newblock In {\em CVPR}, 2013.

\bibitem{zhang2015human}
D.~Zhang and M.~Shah.
\newblock Human pose estimation in videos.
\newblock In {\em ICCV}, 2015.

\bibitem{zhang2013actemes}
W.~Zhang, M.~Zhu, and K.~G. Derpanis.
\newblock From actemes to action: A strongly-supervised representation for
  detailed action understanding.
\newblock In {\em ICCV}, 2013.

\bibitem{zhou2014spatio}
F.~Zhou and F.~D. la~Torre.
\newblock Spatio-temporal matching for human detection in video.
\newblock In {\em ECCV}, 2014.

\bibitem{zhou20153d}
X.~Zhou, S.~Leonardos, X.~Hu, and K.~Daniilidis.
\newblock 3{D} shape estimation from 2{D} landmarks: {A} convex relaxation
  approach.
\newblock In {\em CVPR}, 2015.

\bibitem{zhou2015sparse}
X.~Zhou, M.~Zhu, S.~Leonardos, and K.~Daniilidis.
\newblock Sparse representation for 3{D} shape estimation: A convex relaxation
  approach.
\newblock {\em arXiv preprint arXiv:1509.04309}, 2015.

\bibitem{zhu2014complex}
Y.~Zhu, D.~Huang, F.~{De la Torre}, and S.~Lucey.
\newblock Complex non-rigid motion {3D} reconstruction by union of subspaces.
\newblock In {\em CVPR}, 2014.

\end{thebibliography}

\end{document}


\title{Sparseness Meets Deepness: \\ 3D Human Pose Estimation from Monocular Video \\
\vspace{0.5em}{Supplementary Material}}
\author{}
\date{}

\maketitle

\section*{Proof of Equation (14)}
For simplicity, $\mathcal{L}(\theta;\bfW)$ is denoted as
\begin{align}
\mathcal{L}(\theta;\bfW) & = \frac{\nu}{2}\sum_{t=1}^{n}\left\| \bfW_t - \bfR_t\sum_{i=1}^{k} c_{it}\bfB_i - \bfT_t\bfone^T \right\|_F^2 \nonumber \\
& = \frac{\nu}{2} \left\| \bfW-\bfZ(\theta) \right\|_F^2,
\end{align}
where $\bfW$ is the stack of all $\bfW_t$ and $\bfZ(\theta)$ is the stack of all $\bfR_t\sum_{i=1}^{k} c_{it}\bfB_i-\bfT_t\bfone^T$. Then
\begin{align}
\int \mathcal{L}(\theta;\bfW) \pr(\bfW|\bfI,\theta') d\bfW & =  \frac{\nu}{2} \int \left\|\bfW-\bfZ(\theta)\right\|_F^2 ~ \pr(\bfW|\bfI,\theta') d\bfW \nonumber \\
& =  \frac{\nu}{2} \int \left\{ \left<\bfW,\bfW\right> - \left<\bfW,\bfZ(\theta)\right> + \left<\bfZ(\theta),\bfZ(\theta)\right> \right\} ~ \pr(\bfW|\bfI,\theta') d\bfW \nonumber \\
& =  \frac{\nu}{2} ~ \left\{ \mbox{const} - \int \left<\bfW,\bfZ(\theta)\right> \pr(\bfW|\bfI,\theta') d\bfW + \left<\bfZ(\theta),\bfZ(\theta)\right>  \right\} ~  \nonumber \\
& =  \frac{\nu}{2} ~ \left\{ \mbox{const} - \left< \int\bfW\pr(\bfW|\bfI,\theta')d\bfW~,~\bfZ(\theta)\right>  + \left<\bfZ(\theta),\bfZ(\theta)\right>  \right\} ~  \nonumber \\
& = \frac{\nu}{2}  \left\|\int\bfW\pr(\bfW|\bfI,\theta')d\bfW-\bfZ(\theta)\right\|_F^2 + \mbox{const} \nonumber \\
& = \frac{\nu}{2}  \left\|\mathrm{E}\left[\bfW|\bfI,\theta'\right]-\bfZ(\theta)\right\|_F^2 + \mbox{const}
\end{align}

\section*{Derivation of Equation (15)}

\begin{align}
\mathrm{E}\left[\bfW|\bfI,\theta'\right]
& = \int \pr(\bfW|\bfI,\theta')~ \bfW ~ d\bfW \nonumber \\
& = \int \frac{\pr(\bfW,\bfI,\theta')}{\pr(\bfI,\theta')} ~ \bfW ~ d\bfW \nonumber \\
& = \int \frac{\pr(\bfI|\bfW)\pr(\bfW|\theta')\pr(\theta')}{\pr(\bfI|\theta')\pr(\theta')} ~ \bfW ~ d\bfW \nonumber \\
& = \int \frac{\pr(\bfI|\bfW)\pr(\bfW|\theta')}{Z} ~ \bfW ~ d\bfW \nonumber \\
\end{align}

\section*{Evaluation on HumanEva dataset}

The evaluation results on the HumanEva I dataset \cite{sigal2010humaneva} are presented. The evaluation protocol described in \cite{simo2013joint} was adopted. The walking and jogging sequences from camera C1 of all subjects were used for evaluation. The CNN joint detectors trained on the Human3.6M dataset were fine-tuned with the training sequences for each action separately. Each estimated 3D pose was aligned to the ground truth with the procrustes method. The mean 3D joint errors for the evaluation sequences were reported in \refTab{tab:humaneva}.

\begin{table}[H]
\centering
\renewcommand{\arraystretch}{1.2}
\begin{tabular}{|l|*{3}{c}|*{3}{c}|}
\hline
& \multicolumn{3}{|c|}{Walking} & \multicolumn{3}{|c|}{Jogging} \\
& S1 & S2 & S3 & S1 & S2 & S3 \\
\hline
Proposed & 34.2 & 30.9 & 49.1 & 47.6 & 33.0 & 29.7 \\
Simo-Serra et al. \cite{simo2013joint} & 65.1 & 48.6 & 73.5 & 74.2 & 46.6 & 32.2 \\
\hline
\end{tabular}
\vspace{0.25em}
\caption{Quantitative results on the HumanEva I dataset \cite{sigal2010humaneva}. The numbers are the mean per joint errors in millimeters. }\label{tab:humaneva}
\end{table}


\bibliographystyle{ieee}
\bibliography{bibref_definitions_short,bibref}